\documentclass[review]{elsarticle}

\usepackage{lineno,hyperref}
\modulolinenumbers[5]

\usepackage[margin=1in]{geometry}
\usepackage{float}
\usepackage{graphicx}
\usepackage{subcaption}
\usepackage{siunitx}
\usepackage{amsmath}
\usepackage{makecell}
\usepackage{xcolor}

\newcommand{\retouch}[1]{\textcolor{black}{#1}}
\newcommand{\yuchen}[1]{\textcolor{black}{#1}}
\usepackage[normalem]{ulem}

\begin{document}

\begin{frontmatter}

\title{Simulation Pipeline for Traffic Evacuation in Urban Areas and Emergency Traffic Management Policy Improvements \yuchen{through Case Studies}}





\author[firstaddress]{Yu Chen\fnref{myfootnote}}
\corref{mycorrespondingauthor}
\cortext[mycorrespondingauthor]{Corresponding author}
\ead{yuc2@andrew.cmu.edu}
\fntext[myfootnote]{Work was done while the author was at Google Research.}
\address[firstaddress]{Carnegie Mellon University \\ 5000 Forbes Avenue \\
Pittsburgh, PA 15213}

\author[mysecondaryaddress]{S. Yusef Shafi} 
\author[mysecondaryaddress]{Yi-fan Chen}

\address[mysecondaryaddress]{Google Research \\ 1600 Amphitheatre Pkwy\\  Mountain View, CA 94043}

\begin{abstract}
Traffic evacuation plays a critical role in saving lives in devastating disasters such as hurricanes, wildfires, floods, etc. An ability to evaluate evacuation plans in advance for these rare events, including identifying traffic flow bottlenecks, improving traffic management policies, and understanding the robustness of the traffic management policy are critical for emergency management. Given the rareness of such events and the corresponding lack of real data, traffic simulation provides a flexible and versatile approach for such scenarios, and furthermore allows dynamic interaction with the simulated evacuation. In this paper, we build a traffic simulation pipeline to explore the above problems, covering many aspects of evacuation, including map creation, demand generation, vehicle behavior, bottleneck identification, traffic management policy improvement, and results analysis. We apply the pipeline to two cases studies in California. The first is Paradise, which was destroyed by a large wildfire in 2018 and experienced catastrophic traffic jams during the evacuation. The second is Mill Valley, which has high risk of wildfire and potential traffic issues since the city is situated in a narrow valley.
\end{abstract}


\end{frontmatter}
\section{Introduction}
Evacuation in urban areas plays a key role in protecting and saving lives during natural disasters and hazards, such as wildfire, hurricanes, bomb threats, etc. Insufficient evacuation route throughput leads to traffic jams and gridlock, traps vehicles on roads, and leaves residents in danger during emergencies. In the 2018 Camp Fire in Paradise, California, many drivers had to abandon their cars and flee on foot while surrounded by fire \cite{paradise2018timeline}. Similar problems were also seen for hurricanes \cite{hurrican2017jam,hurrican2017floridajam}, bomb threats \cite{bomb2012jam}, and tsunamis \cite{tsunami2013jam}. It is critical to analyze evacuation plans and to propose efficient operational strategies before tragedy strikes. Several fundamental questions must be addressed, such as ``how long does it take to evacuate the area?", ``where is the bottleneck of the network during the evacuation?", ``is there a better solution?", and ``is there a flexible platform to test all kinds of evacuation plans?". Our work explores the answers for these questions by building a platform and a pipeline for the evacuation simulation, and by presenting two case studies of whole city evacuation of Paradise and Mill Valley, two cities located in wildfire-prone regions in California, using microscopic traffic simulation.

Historical traffic data of catastrophic events is usually not available. Development of new traffic management policies further necessitates analyzing the consequence of proposed changes in the absence of observed data. \yuchen{ Analytical methods for traffic analysis are typically based on very detailed knowledge of traffic flow and assumptions that are in many cases inherently challenging to estimate and validate, respectively, making simulation a viable alternative that can sidestep some of these restrictions. In these situations, traffic simulation provides a convenient way to empirically study traffic policies.} Traffic simulation tools can be generally classified into microscopic, mesoscopic, and macroscopic models \cite{kessels2019traffic}. Microscopic models simulate individual cars by calculating each vehicle's speed and location using car-following and lane-changing modules. Macroscopic models depict the traffic stream as continuum flows and apply equations similar to those in fluid dynamics to describe the traffic density dynamics, traffic speed and traffic flow. This simplification gains a lot of benefits in computation, control and system identification. Mesoscopic models are in between the above two approaches, grouping the transportation elements into clusters, describing vehicle flow in terms of probability distributions, and simplifying the traffic flow using queues or similar methods \cite{kessels2019traffic,krajzewicz2012recent}. We choose to use an open source microscopic simulator called Simulation of Urban Mobility (SUMO), since it is realistic and provides detailed interfaces for drivers' behaviors such as lane changing and car following \cite{krajzewicz2012recent}. SUMO also includes a suite of tools for simulation, such as a map editor, trips generator, data extraction scripts, etc. For our case studies, behavioral realism is of key importance, so we choose a more detailed simulator rather than one with faster computation models.

Usually the detailed records of vehicles' tracks from previous evacuation are not available, and because of driver behavior variability, distribution of vehicles, road network conditions, etc., it is challenging to use a fixed model to cover the broad general cases. A main principle of our study is to create a sequence of scenarios with different parameters instead of using one set of parameters. We apply this principle in creating demands, routing, and network conditions. For example, the concentration of the temporal distribution of the demands spans from narrow to wide, representing the \yuchen{no-notice urgent and short-notice moderate evacuation modes, respectively}. The portion of the total number of vehicles needing to be evacuated ranges from 50\% to 100\%, representing different traffic volumes on the road network.

One remarkable advantage of traffic simulation is that it allows researchers to modify arbitrarily any aspects of the evacuation process with almost no cost, including driver behavior models, traffic management policies, the road network, and trip demands. In addition, the microscopic simulator yields the detailed trajectory tracks of each individual vehicle. These allow us to accurately identify bottlenecks in the traffic evacuation plan, predict the evacuation time, make improvements, and test traffic policies. Some previous studies improve the evacuation efficiency by optimizing the routing methods \cite{zyryanov2017simulation,shahabi2018scalable,li2014optimization}, adding extra roads \cite{jha2004emergency}, or changing the traffic light cycles  \cite{chen2007traffic,parr2011critical}. These methods highly rely on driver behavior and awareness, infrastructure construction, centralized control, or power supply, and may have a higher risk of failure. Our simulation study shows some intersections are the bottlenecks of the evacuation and they are the core issue limiting the capacity of the road network. This problem is also reported by eyewitnesses of the Paradise Camp Fire \cite{john2018paradise}. Thus improving the intersections is our highest priority in this paper. We seek simple, consistent, robust traffic management policies which require little supervision, public education, changes to infrastructure, power supply, and driving skills.

Contraflow is a promising strategy to solve the intersection problem which is a lane-based manipulation strategy for the evacuation traffic management policy \cite{xie2011lane}. It enlarges the transition capability in outbound directions by reverting, blocking, or reconnecting some lanes. This strategy is attractive since it is easy to deploy and driver-friendly. It also requires minimal supervision, power supply, and infrastructure changes. Previous research and reports show that contraflow is a highly effective strategy in both simulations and real cases \cite{wolshon2001one,urbina2003national,shekhar2006contraflow,meng2008microscopic,tuydes2006tabu,xie2011lane,pyakurel2015models}. We will present the performance of the proposed traffic management policy based on contraflow in our case studies in section \ref{sec:case_studies} and \ref{sec:results}.

In this paper, we provide a pipeline for evacuation traffic simulation, and we have made it publicly available\footnote{The code and traffic case studies are available at \url{https://github.com/google-research/google-research/tree/master/simulation_research}}. It includes maps creation, demands generation, vehicle routing, traffic management policy design, microscopic traffic simulation, data collection and results analysis. Two case studies are presented using our pipeline. We analyze how different components in the pipeline affect the evacuation efficacy. \yuchen{ We hope other researchers or city managers who need to interactively design and test their evacuation strategies can benefit from our tool chain.} The rest of this paper is organized in the following manner. Section \ref{sec:methods} introduces the simulation platform, and the components of an evacuation scenario. Section \ref{sec:case_studies} presents two case studies, and the rationales for the scenarios' chosen parameters. Section \ref{sec:results} shows the results of the scenarios. Section \ref{sec:discussion} provides a discussion of the results. Finally, we conclude the paper with Section \ref{sec:conclusion}.

\section{Simulation pipeline} \label{sec:methods}
Traffic simulation consists of many elements, including maps, vehicle behavior, traffic demands, and results analysis and evaluation tools. We build an end-to-end pipeline for evacuation traffic simulation, which enables us to examine the details of an evacuation simulation vis-a-vis the above facets.

\subsection{Traffic simulator}
We use SUMO version 1.1.0 as the simulator. It is an open source microscopic space-continuous time-discrete traffic simulator \cite{krajzewicz2012recent}. The time step length is 1 second, chosen to provide an empirically reasonable balance between computational complexity and accuracy. The maximum duration of a scenario is 49 days. In our study, the simulator handles tens of thousands vehicles in total. The simulator can run on a single core of a CPU. There are two ways to run the simulation: using \textit{sumo} and \textit{sumo-gui}. The former one is a command line tool, the latter one is a graphical user interface, which can be used to verify the map, debug the scenarios, and visually monitor the traffic flow. Our work simulates around 10 hours of traffic, which takes from 15 minutes to one hour of computation time depending on the scenario. Each vehicle is modeled independently as a particle with some physical properties such as car length, maximum speed, maximum acceleration, etc. A vehicle’s location is described by the relative position and the lane on a segment of road. The vehicle's speed is computed using a car-following model. The current integrated algorithm is introduced by \cite{krauss1998microscopic}. In addition, SUMO incorporates lane changing behaviors. The details of the algorithm can be found in \cite{erdmann2015sumo}. The SUMO suite also provides tools for processing and generating maps and traffic routing. 

\subsection{Maps} \label{subsec:methods_maps}
We obtain open source maps from OpenStreetMap \cite{openstreetmap}. The maps represent physical features using tags in XML files. The map can be queried using ranges of latitude and longitude. A tag in the XML file can be a node, an edge (road), a relation, or an associated geographic attribute. A curvy road is approximated using a sequence of line segments. The roads intersect at nodes, and the relation describes the connections of edges or lanes. The OpenStreetMap maps need to be converted to SUMO-specific XML-representations using \textit{netconvert} \cite{netconvert}. The \textit{netconvert} tool also accepts other popular map types such as VISUM, Vissim, or MATSim. The SUMO XML representation is composed of five elements, including nodes, edges, (optional) edge types, connections, and (fixed) traffic light plans. However, the maps may sometimes have flaws, including incorrect (1) lane connections from freeway ramps to the main travel lanes, (2) stop sign assignments, (3) road intersections with footpaths, (4) number of lanes, (5) road types, priority or speed limit, and (6) traffic lights. These problems are rare, but can be found by visual inspection, simulation, or comparison with mapping software such as Google Maps Street View. A practical heuristic strategy for verifying maps is to examine them from a large scale to a small scale. For example, we first inspect arterial roads by highlighting them in \textit{netedit}. The \textit{netedit} tool is a visual network editor included with the SUMO suite. It can be used to create or to modify road networks. With a powerful highlighting and selection interface, it can easily be used to debug network attributes. Residential roads and local roads can be verified subsequently. This step makes sure the roads are of the correct type, priority and other attributes. Next, we inspect details such as connections of roads and lanes. Issues are less obviously detected by simple highlighting, so we use injected random or specified traffic flow into the network to examine the functions of the connections. Wrong connections, road types, numbers of lanes, traffic lights, or stop signs usually cause blockages or significant slowing, so it is still easy to find issues using the graphical tool \textit{sumo-gui} or an offline speed map generated by the simulation. Another way to check connections is to plot the shortest paths between several selected roads.

\subsection{Vehicles}
As SUMO is a microscopic traffic simulator, each vehicle is modeled explicitly. The attributes of a vehicle include a unique identifier, departure time, departure position, departure speed, destination, and a route for the trip. A full list of physical parameters is in \cite{sumovehicle}. The settings follow the study of \cite{krauss1998microscopic}. The car following model is also based on the work of Krauss  \cite{krauss1998microscopic}. In brief, this model lets vehicles drive as fast as they can without violating safety constraints. The model avoids collision (is collision-free) if the leader starts braking within the maximum deceleration window. In real world, driving speed usually varies within traffic flow. In SUMO, this is modeled by a distribution of speeds parameterized by a mean and standard deviation. The mean speed is set to the lane speed limit (if there is no car in front), and the deviation is set to 0.1 multiplied by the mean value. We justify our uniform vehicle assumption by noting that the majority of both cities in our case studies is residential, and so we assume that types of vehicles such as large trucks or buses constitute a small portion of all vehicles.

\subsection{Demands} \label{sec:demands}
After creating a network and configuring the vehicles, SUMO needs demands files to generate traffic flows. A trip for each individual vehicle needs a starting location, departure time, a destination, and a path. The routing methods will be introduced in the following paragraph. SUMO ingests a demands file prior to running a simulation. The demands file can specify either the trip of individual vehicles or a traffic flow with the same origin-destination pairs \cite{sumodemands}. It can also specify a deterministic path or simply an origin-destination pair with dynamic rerouting \cite{sumorouting}. \yuchen{As for the temporal distribution of demands for evacuation events, we follow the model of many previous works \cite{jha2004emergency,sorensen2000hazard,wolshon2001one}, which agrees with empirical observations of wildfire and hurricane evacuations, \cite{loreto2019temporal,woo2017reconstructing,usarmy2001hurrican,usarmy2003hurrican}. and is also similar to our case study. A rough timeline for Paradise evacuation is described in section \ref{subsec:scenarios_demands}, which has low demands in the beginning and the end, but has a peak in the middle. } The cumulative evacuation vehicles follows an S-curve. There is a delay between the onset of the event and the peak of the traffic flow due to the latency of the evacuation notification. There are relative smaller traffic flows before and after the peak. If the alert is sent out with a no-notice, it will trigger sharper demands with a steeper S-curve. On the other hand, if the alert does not reach the full targeted population at the same time, or there is no centralized alert at all, the demands will be distributed over a larger time range with a flatter S-curve. In addition to the timing of the vehicle generation, the simulation also needs to know where and how many vehicles need to be created, and where they will go. In the evacuation case, the origins of the vehicles can be anywhere in a city, including residential and commercial areas. The destinations are usually limited, such as freeway or arterial road exits of the city \cite{jha2004emergency,zimmerman2007using}. Some scenarios may also consider shelters \cite{cova2011modeling}. If the affected area is small, vehicle trips only need to avoid the danger area, and can still preserve their original destinations \cite{flotterod2018dynamic}. Usually a city has multiple exits, and taking any one of them is equivalent from a safety perspective. Instead of fixing a destination road for each vehicle, the simulation allows grouping several roads into a traffic analysis zone (TAZ) as the destination. If the destination is set as a TAZ, the automatic routing algorithm will chose the closest road among the TAZ \cite{sumoautorouting}. Details for the spatial and temporal distributions of the demands in our case studies will be discussed in the next section.

\subsection{Routing methods} \label{sec:routing}
The route of a trip in SUMO is composed of a complete sequence of adjacent roads between the origin and the destination. The simulator allows adding more details about the departure and arrival properties, such as the lanes, the velocity, or the position on an edge. To be more realistic, the route can also be determined dynamically according to the traffic situation using automatic rerouting \cite{sumoautorouting}. This method equips all or part of the vehicles with the ability to update their optimal paths periodically. It considers the current and recent state of traffic in the network and thus adjusts the plan to traffic jams and other changes in the road network. If the automatic rerouting function is enabled for a vehicle, the average travel times are computed for the roads in the network. The vehicle will choose the fastest path to its destination according to the roads' mean travel speeds. A road's mean travel speed or travel time is updated periodically by using not just the most recent record, but instead an average of the past few records. In this way, the routing method can avoid shortsighted plans and be more consistent. Periodically updating the network status and planning is a heuristic way to approximate the global dynamic users' equilibrium. However, \cite{lindell2007critical,pel2012review} criticize the use of user equilibrium for evacuations as it is difficult for the evacuees to learn traffic conditions in rare events and traffic rules can differ from their usual state. In our case studies, the road networks are tree-like graphs, so path planning is less undetermined. Although SUMO can simulate the drivers' behavior relatively close to reality, there are still some limitations on mimicking flexible movements. For example, if two cars from different directions are both trying to make left turns at the same intersection, the vehicles might get stuck. To avoid deadlocks, \textit{teleporting} is used \cite{sumoteleport}. In the case that vehicles at the first position in front of an intersection have waited longer than 300 seconds (or an alterante chosen threshold), the vehicle is assumed to be in gridlock and is teleported. A teleported vehicle is temporarily taken away from the network, and then virtually moved forward along its route. It is inserted into the network as soon as this becomes physically possible.

\subsection{\yuchen{Network disruption or road capacity reduction}}
The condition of the road network can be disturbed by disasters such as floods, hurricanes, and wildfires. We are interested in exploring the consequences of disaster-induced road blockages, especially that of arterial roads during evacuation, through simulation. SUMO allows a certain road to be blocked dynamically during a specific time period \cite{sumorerouter}. This can be used to evaluate the robustness of the traffic management policy. If a vehicle has a deterministic, fixed routing path to the blocked exit, it will get stuck on the road in the simulation. If the vehicle is equipped with automatic rerouting, and there is a path to another exit when the previous exit is blocked, it will change its path.

\subsection{Simulation outputs and data analysis}
SUMO records the positions of all vehicles every second through the simulation. The time step can be smaller, but to balance the trade-off between accuracy and computation, we choose to use one second as the time step size. The output data is summarized in different forms. Floating car data (FCD) files export locations and speeds along with other attributes for every vehicle at every time step \cite{sumooutput}. The FCD file can be used to trace the trajectories of the vehicles along with the traffic conditions. The FCD can be thought as records from high-accuracy and high-frequency GPS devices on vehicles. A speed map is created by averaging the speeds of the passing vehicles every half hour for each road. This can be used to understand the dynamics of the evacuation and to identify bottlenecks where the speed is particularly slow. As the simulation is recorded in detail for every vehicle, the FCD file size is quite large, on the order of several GB. The summary file contains high-level information at a coarser granularity. It has the complete simulation-wide number of vehicles that have been loaded, inserted, are running, and have finished. The cumulative counts or percentages of finished vehicle trips can be used to describe the evacuation process and to measure the evacuation efficacy. Other types of the output files can be used as supplementary data. Induction loop detectors can be embedded within a given lane of a road. They count the number of vehicles that pass over them and the mean velocity of all counted vehicles. The detectors monitor the traffic at a fixed location, rather than the trajectories of individual vehicles. The evacuation efficacy is quantified as the gap area between the cumulative curve of the demands and that of the number of evacuated vehicles in time. It is calculated as the difference between the areas under the curve. \yuchen{For example, in Figure \ref{fig:demands_evacuation_curves} (a), the gap area between the demands and one evacuation curve (shown in green) is shadowed in gray. The gap area can be understood as the average trip time if the area is seen as the Riemann integral along the percentage axis: }
\begin{equation}
\text{gap area} := \sum_i (\text{percentage}_{i+1} - \text{percentage}_i) 
\Big( f_{\text{demands}}(\text{percentage}_i) - f_{\text{finished}}(\text{percentage}_i) \Big)
\label{eq:gap_area}
\end{equation}
where $\text{percentage}_i$ is the percentage value at interval $i$, the function $f_{\text{demands}}$ or $f_{\text{finished}}$ is the inverse function of the cumulative function. The difference $ f_{\text{demands}}(\text{percentage}_i) - f_{\text{finished}}(\text{percentage}_i)$ represents the time gap between the demands curve and the evacuation at $\text{percentage}_i$.

\section{Simulation case studies} \label{sec:case_studies}

We present case studies about full-scale traffic evacuation of the whole city under destructive, life-threatening emergencies, such as a wildfire or hurricane. Regional evacuation in small areas is not the main focus of this work. In order to analyze scenarios with short-notice or no-notice evacuations, we create a sequence of demands with more or less concentrated temporal distributions. We aim to predict the evacuation time duration to completion, identify the bottlenecks, and ascertain the reasons for observed traffic congestion. If the current traffic management policy cannot handle the evacuation efficiently, we seek improvements involving minimal changes to the infrastructure or simple-to-implement policies. The robustness of the traffic management policy is another important aspect of the evaluation as natural disasters may perturb the network by blocking or slowing down roads, and large traffic flow volumes may cause increased car collisions.

\begin{figure}[H]
\centering
\begin{subfigure}[t]{1\textwidth}
    \centering
    \includegraphics[height=10cm]{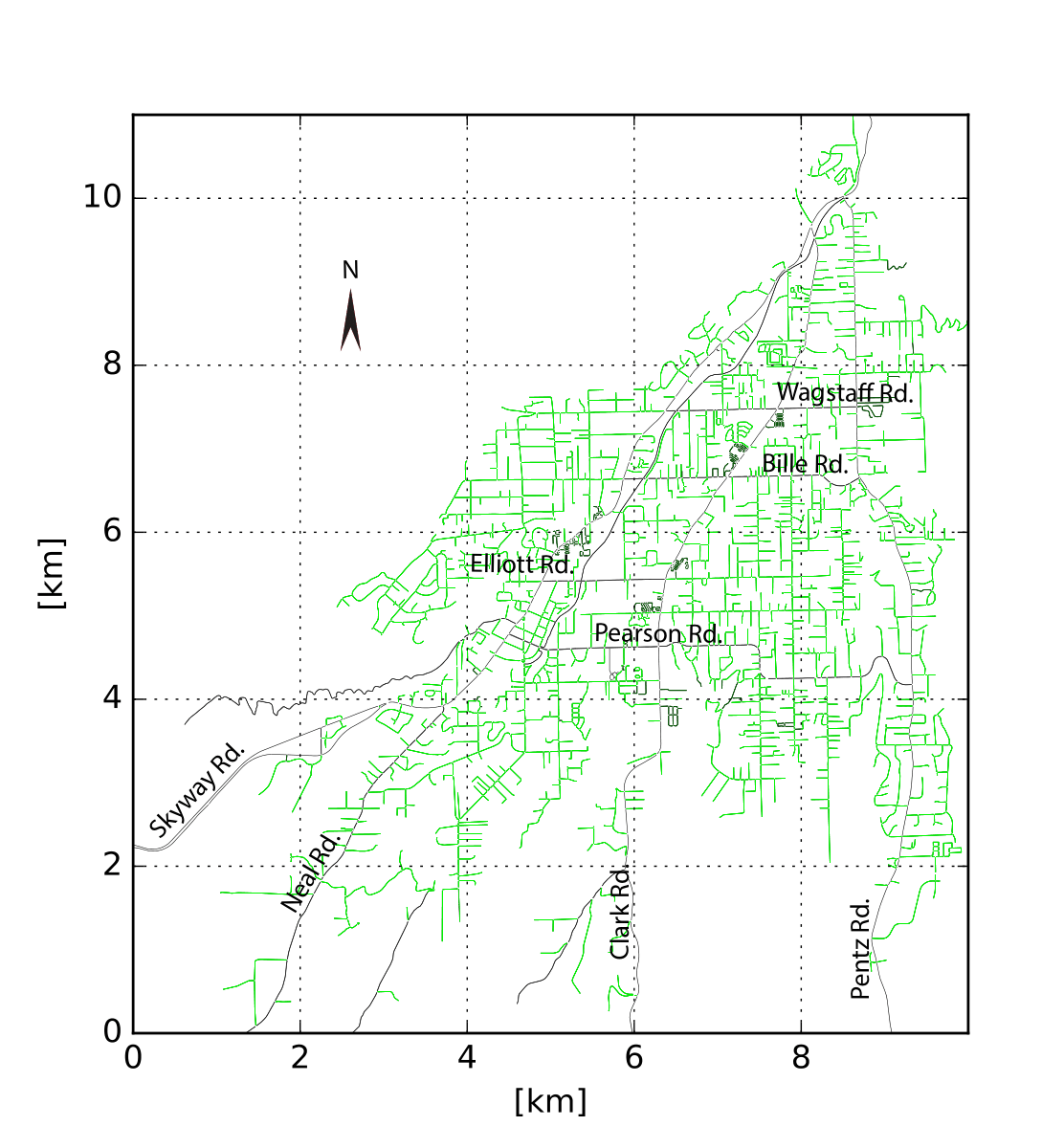}
    \caption{Paradise, CA U.S.}
\end{subfigure}%

\begin{subfigure}[t]{1\textwidth}
    \centering
    \includegraphics[height=10cm]{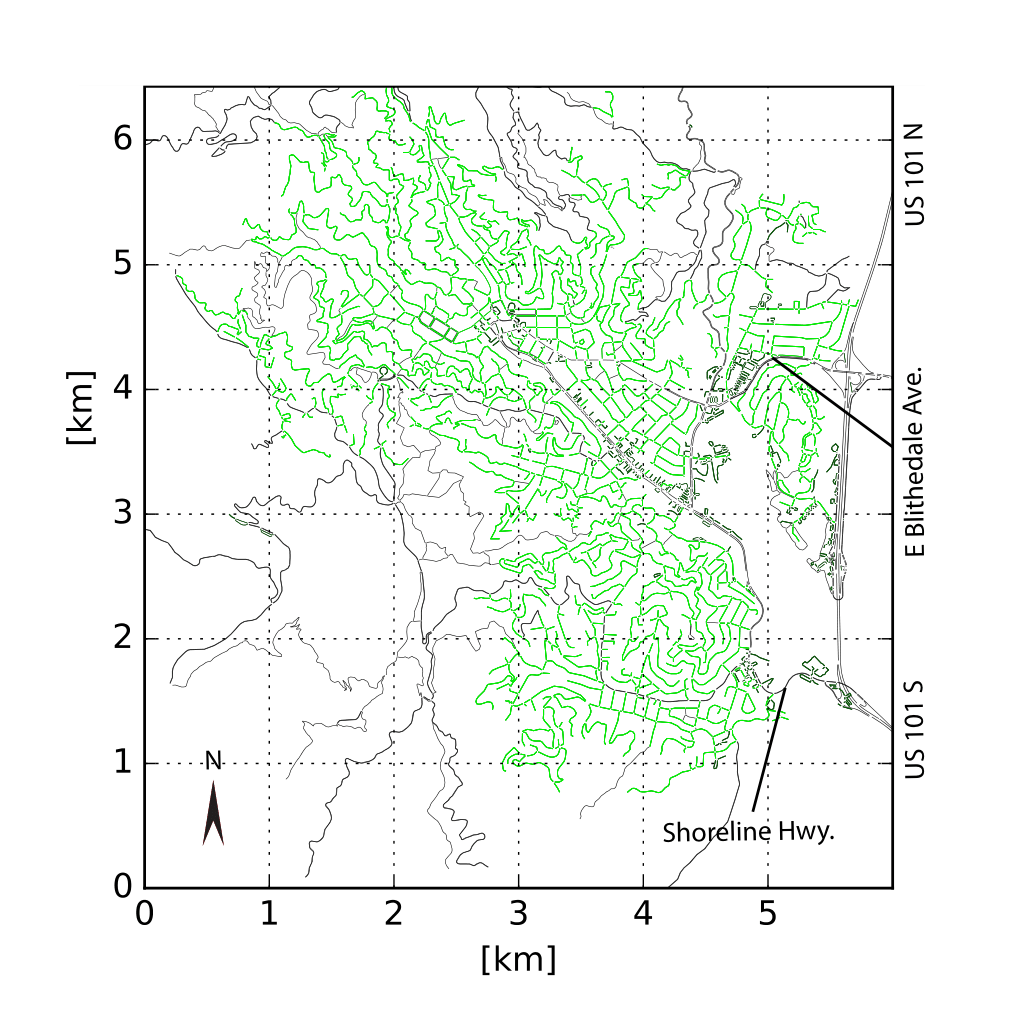}
    \caption{Mill Valley, CA U.S.}
\end{subfigure}%

\caption{The maps of Paradise and Mill Valley. Arterial roads are shown in black, residential roads are shown in light green and parking areas are shown in dark green. (a) Paradise. The evacuation exits are Skyway Rd., Neal Rd., Clark Rd., and Pentz Rd. Neal Rd. is set as the rescue entrance. (b) Mill Valley has four exits in two directions, including two exits through E Blithedale Ave. and two exits through Shoreline Hwy.  Shoreline Hwy. is set as the rescue entrance.}
\label{fig:maps}
\end{figure}

Evacuation traffic simulations of two cities in California with demonstrated or experienced wildfire risk are presented for Paradise and Mill Valley. Paradise experienced a severe wildfire on November 8, 2018. This fire was known as the deadliest and most destructive wildfire in California history \cite{baldassari2018campfire}. It was the world’s costliest single natural disaster in 2018, which caused 86 fatalities, the loss of 13,900 homes \cite{john2018paradise}, and overall losses of 16.5 billion US dollars \cite{time2019campfire}, destroying the majority of the city. During the evacuation, large outbound traffic flows led to intersections being quickly jammed, trapping and endangering hundreds of residents in the city. It was reported that some residents tried to flee on foot and abandoned their cars as they were stalled in traffic jams \cite{nytimes2018paradise}. The fire started seven miles to the east of the city, and expanded quickly to the Pentz Road, the arterial road on east side of the city \cite{john2018paradise} (also see Figure \ref{fig:maps} (a)). Before the disaster, the city officials did not know precisely how much time was needed to evacuate the city \cite{john2018paradise}.

\retouch{At the state level, roughly ten percent, or more than 1.1 million structures, lie within the highest-risk fire zones as determined by the California Department of Forestry and Fire Protection \cite{latimes18}. A significant portion of the City of Mill Valley lies in a very high fire hazard zone, as determined by the state \cite{calfire}.}

Microscopic traffic simulation provides an opportunity to study the evacuation by examining many aspects, including maps, demands, routing, traffic management policy and the robustness of the evacuation plan. Usually the detailed historical data for traffic evacuation are not available, especially for the rare events. Instead of making very narrow and specific assumptions, we perform simulations across a broad range of scenarios. By pursuing a risk-based approach to simulation, the distribution of outcomes provides coverage of a range of realistic scenarios, for example, from no-notice to short-notice, from larger to smaller flows of traffic, or from worse routing strategies to closer to optimal ones. The details and the rationale are given as follows. The techniques about these aspects are discussed in section \ref{sec:methods}. We summarize the scenarios in table \ref{tab:scenarios}.

\begin{table}[H]
\centering
\def\arraystretch{1.2}
\begin{tabular}{|c|c|c|c|c|c|}
\hline 
Scenario & City  & \makecell{Demands distribution \\ \yuchen{standard deviation} [hour]} 
& \makecell{Portion of \\ total vehicles[\%] } & Routing & \makecell{Traffic  rules}  \\ \hline
1 & Paradise    &  0.7  &  100  & Shortest path & Normal    \\ \hline
2 & Paradise    &  0.7  &  100  & Auto-rerouting &  Normal  \\ \hline
3 & Paradise    &  0.7   &  100  & Auto-rerouting & Contraflow \\ \hline
4 & Paradise    &  0.2 -- 1.5   &  100  & Auto-rerouting & Contraflow \\ \hline
5 & Paradise    &  0.2   & 50 -- 100  & Auto-rerouting & Contraflow \\ \hline
6 & Paradise    &  0.7   & 50 -- 100  & Auto-rerouting & Contraflow \\ \hline
7 & Paradise    &  0.7   & 100  & Auto-rerouting & Road block \\ \hline
8 & Mill Valley  &  0.5  &  100  & Shortest path & Normal  \\ \hline
9 & Mill Valley  &  0.5  &  100  & Auto-rerouting & Normal  \\ \hline
10 & Mill Valley  &  0.5  &  100  & Auto-rerouting & Contraflow  \\ \hline
11 & Mill Valley  & 0.3 -- 1.0 &  100  & Auto-rerouting & Contraflow \\ \hline
12 & Mill Valley  & 0.5 &  50 -- 100  & Auto-rerouting & Contraflow  \\ \hline
13 & Mill Valley  & 0.7 &  50 -- 100  & Auto-rerouting & Contraflow \\ \hline
14 & Mill Valley  & 0.5 &  100  & Auto-rerouting & Road block  \\ \hline
\end{tabular}%
\caption{All simulation scenarios for the case studies. We consider how demands distribution (section \ref{subsec:scenarios_demands}),  portion of total vehicles (section \ref{subsec:scenarios_demands}), routing methods (section \ref{subsec:scenarios_routing}), traffic management rules (section \ref{subsec:traffic_management_policy}) and road network disruption or road capacity reduction. (section \ref{subsec:scenarios_robustness}) influence the efficient of the evacuation. }
\label{tab:scenarios}
\end{table}

\subsection{Maps}

The maps are acquired from OpenStreetMap \cite{openstreetmap} by querying relevant ranges of latitude and longitude. For Paradise, the range of the map is in the box bounded by \ang{-121.779150} (west bound), \ang{39.602528} (south bound), \ang{-121.551501} (east bound), and \ang{39.841422} (north bound). For Mill Valley, the range of the map is in the box bounded by \ang{-122.596980} (west bound), \ang{37.859704} (south bound), \ang{-122.485271} (east bound), and \ang{37.935805} (north bound). See maps in Figure \ref{fig:maps}. Some roads in corners of the maps that do not belong to the cities are removed. The map of Mill Valley extends to Tamalpais Valley and Almonte on the south side of the city because these areas are adjoined and share the exit arterial roads, so we include them to take account of the traffic interactions. Flaws of the map are corrected following the methods outlined in section \ref{subsec:methods_maps}. We aim to design evacuation policies with minimal change of the existing infrastructure, so extra roads are not added to the maps.

\subsection{Demands} \label{subsec:scenarios_demands}

\begin{figure}[ht]
\centering
\begin{subfigure}[t]{0.5\textwidth}
\centering
\includegraphics[height=6cm]{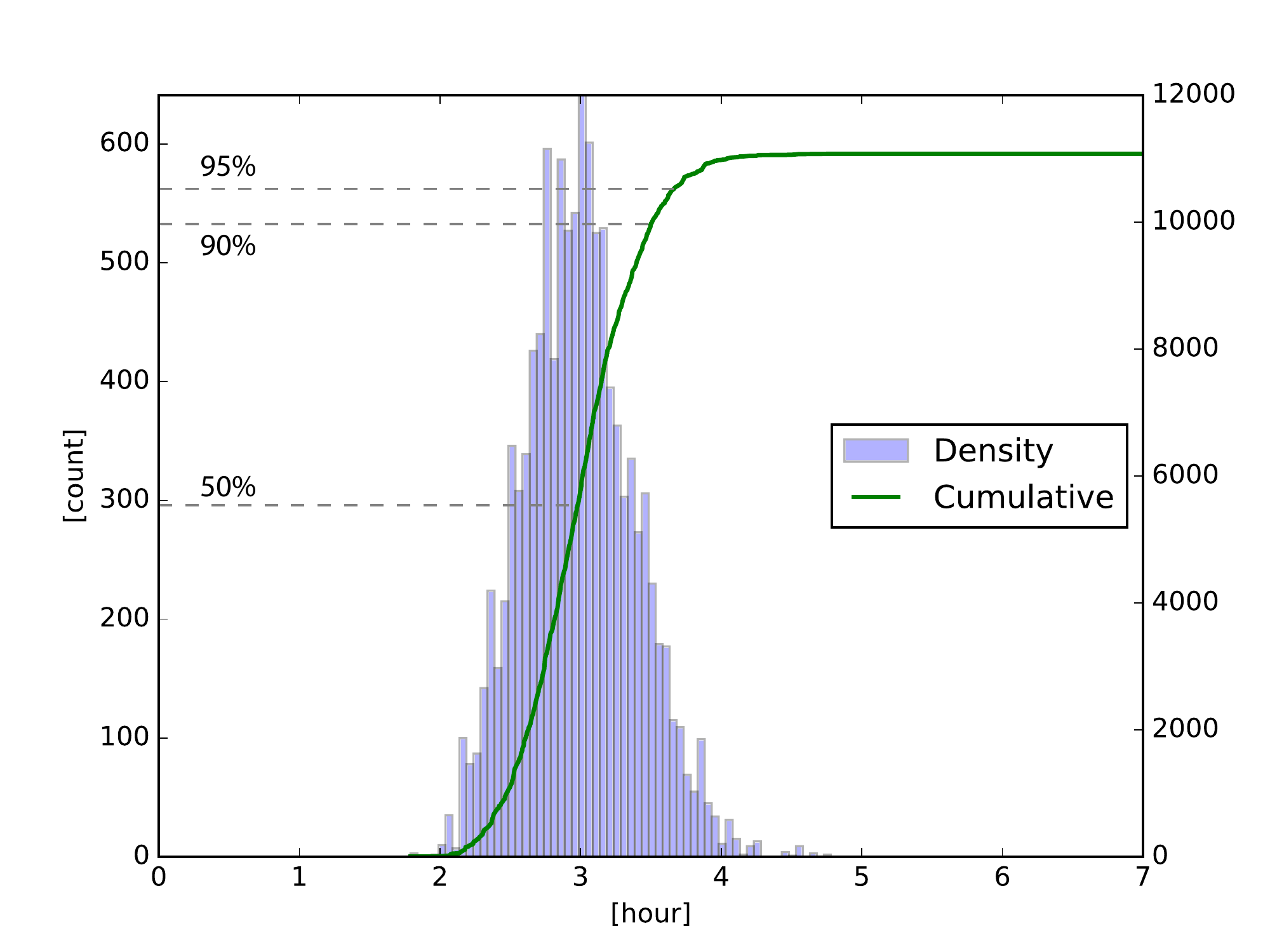}
\caption{Demands with high concentration.}
\end{subfigure}%
\begin{subfigure}[t]{0.5\textwidth}
\centering
\includegraphics[height=6cm]{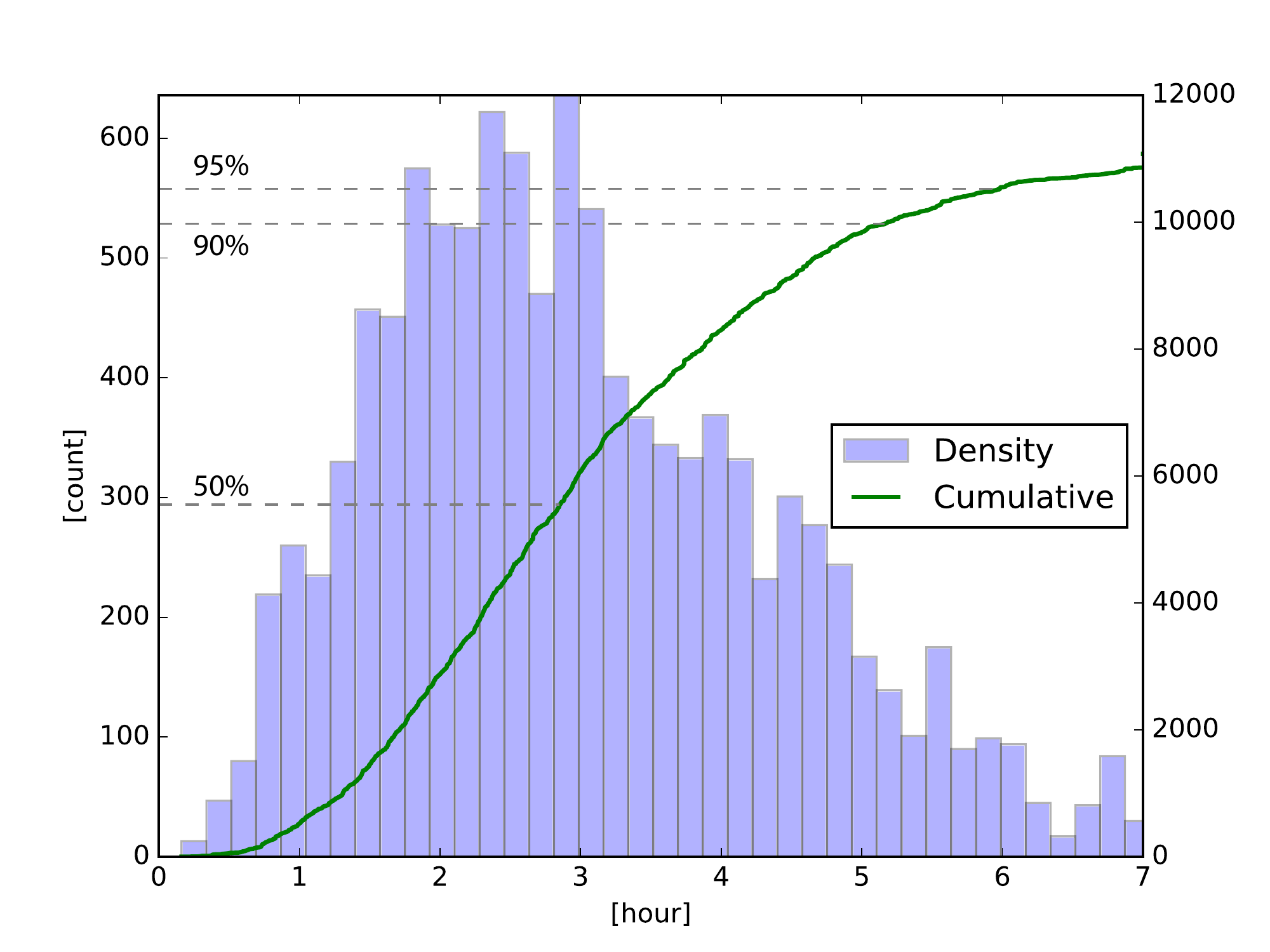}
\caption{Demands with low concentration.}
\end{subfigure}%

\caption{Two examples of demands temporal distribution with high and low concentration, and the corresponding cumulative curves. The departure times follow a Gamma distribution. The means are 3.0 hours, and the standard deviations are 0.4 hours in Figure (a) and 1.5 hours in Figure (b). Figure (a) represents the more urgent evacuation or no-notice. Figure (b) represents less urgent evacuation or short-notice. The vertical lines indicate 50\%, 90\% and 95\% of the total demands. Other distributions have the same mean value, but with different standard deviations (figures are not shown). }
\label{fig:demands}
\end{figure}

\retouch{Generally speaking, traffic demands for evacuations must take into account spatial and temporal aspects. All generated vehicles must leave the city via the exit roads. This work studies stimulation of large scale evacuation under severe disasters, so localized evacuation of small regions and shelter-in-place plans are not considered. In the catastrophic situations hypothesized in this paper, shelters are assumed to be unavailable or unable to provide sufficient protection against disasters, and as such evacuation is required \cite{xie2011lane}. In the case of cites with dense residential areas and proximity to wildlands, partial evacuation may be challenging in the presence of widespread fires. Furthermore, fire prevention plans to control wildfire spread may be insufficient in high-wind situations \cite{fireprevention2019}. In this paper, we consider hypothetical worst case scenarios, i.e., the evacuation of the whole population. Our model assumes the origin positions of vehicles are distributed uniformly along residential roads, and that their temporal distribution follows a bell-shaped curve as introduced in section \ref{sec:demands}.}

For the spatial distributions of demands, after examining the Google Maps satellite imagery of both Paradise and Mill Valley, we find that the majority consists of residential areas with single-family houses. The houses are located on both sides of the residential roads with roughly equal spacing. There are also a few large parking areas near grocery stores, churches, and other community facilities. All the vehicles are set to start in residential roads or service roads, not in arterial roads or freeways. According to Data USA \cite{datausa} and 2010 United States Census \cite{cencus2010}, Paradise has 10,700 households, and Mill Valley has 6,000 households. On average, each household has two cars. The total lengths of the residential roads in the simulation maps are 514 km and 337 km for Paradise and Mill Valley, respectively. The total lengths of service roads are 26 km and 42 km, respectively. The length of the road is calculated using OpenStreetMap data \cite{openstreetmap}. Residential roads are shown in light green in Figure \ref{fig:maps}, and service roads are shown in dark green. Thus we can determine the average vehicle density, which is the number of vehicles per unit length of road. We assume the vehicle density in parking ares is larger than that of the residential roads, the ratio is set as four. There are roughly four cars within a house-size road in the satellite images. It is hard to estimate this ratio, so we choose the number arbitrarily. We test with different ratios, they do not affect the conclusions. In Paradise, the vehicle density for residential roads is 0.0417 vehicles/meter and for service roads is 0.1668 vehicles/meter. In Mill Valley, the vehicle density for residential roads is 0.0270 vehicles/meter and for service roads is 0.1080 vehicles/meter. The roads with length smaller than 24 m in Paradise and 37 m in Mill Valley are not included in the calculation, as the number of vehicles on the road is smaller than one using the above calculation. The total number of simulated vehicles is 23,635 for Paradise and 12,212 for Mill Valley. The exits of Paradise and Mill Valley are shown in Figure \ref{fig:maps}. Those exit roads are grouped into a single TAZ for the purposes of automatic rerouting. In section \ref{subsec:scenarios_routing}, we introduce more efficient evacuation plans by reversing inbound roads. However, those plans still keep at least one entrance road open for rescuing school students, the elderly, persons with disabilities, and people who cannot drive. The exits are selected with the lightest burden so they will not obstruct the main efflux traffic stream. In Paradise, the Neal Rd. is a branch of Skyway Rd. and it covers a relatively smaller area than others as illustrated in Figure \ref{fig:maps}. Similarly, the Shoreline Hwy. is kept as entrance for Mill Valley.

For the temporal distribution of the demands, we make use of the S-curve models commonly seen in the evacuation literature \cite{jha2004emergency,sorensen2000hazard}. The empirical observations also agree with the model \cite{usarmy2001hurrican,usarmy2003hurrican}. Thus we use a Gamma distribution for vehicles departure times as shown in Figure \ref{fig:demands}. The exact ground truth of the evacuation timeline is not available, but there are still some clues from the news about the Paradise Camp Fire supporting the proposed model \cite{paradise2018timeline}. The wildfire started 6:29 AM at Pulga, CA, seven miles away from Paradise. At 7:23 AM, the
Butte County Sheriff tweeted an evacuation order for Pulga. At 8:03 AM, the Butte County Sheriff tweeted the first evacuation order for Paradise. At 9:17 AM, a full-scale evacuation order was issued, but by then the fire was already consuming the town. Heavy traffic lasted for hours after the order \cite{paradise2018timeline}. Thus the mean of the distribution, which is very close to the peak in the Gamma distribution, is set to three hours following the evacuation. Different cities may have different evacuation alert systems, and to obtain more general conclusions from the study, we are more interested in studying the gap times between the initial demand and the end of the evacuation. Instead of fixing the standard deviation of the demands, we set it in a wide-enough range to cover no-notice and short-notice cases. We propose a more advanced model in section \ref{sec:discussion} for more complicated spatially- and temporally-dependent demand models. In table \ref{tab:scenarios}, the column ``Demands distribution" indicates the temporal distributions for different scenarios. 

Next, we study whether reducing the number of vehicles can increase the evacuation efficacy. If this is an efficient solution, the residents should be encouraged to take carpools, footpaths, or other methods to reduce the total number of departing vehicles. For example, Mill Valley has many footpaths covering a large portion of hillside residential areas. We present the simulation with partial vehicle utilization ranging from 50\% to 100\%. In table \ref{tab:scenarios}, the column labeled ``Portion of total vehicles" indicates the portions for different scenarios. Our work does not explicitly model the background traffic in the city area or along the exit roads. Different portions of the total vehicles can include more or less background traffic within the city area. For the background traffic along the exit roads, as long as the efflux traffic is not blocked, it is reasonable to simplify the model. The case of exit roads themselves becoming bottlenecks due to external background traffic jams outside of the city jurisdiction is out of the scope of this paper. We discuss the traffic analysis with respect to external factors in section \ref{sec:discussion}. We point out potential issues regarding persuading the public to reduce the number of cars by leaving their properties behind in section \ref{sec:discussion}.

\subsection{Routing} \label{subsec:scenarios_routing}
Modeling driver behavior is complex and challenging in general, as it includes many human or non-human factors with large variance, not to mention changes during life threatening events due to driver anxiety and the destruction of transportation and communication infrastructure. Those factors related to driver behavior include a driver's previous experience of hazards, environmental cues, receiving a warning, injuries present, distance to the hazard, access to social networks, children in the house, gender, age, fear of looting, \yuchen{and} others  \cite{murray2013evacuation}. Collecting the data to model these factors is non-trivial and the model cannot be easily generalized to other cases. So instead of considering one specific model, we present models with different parameters ranging from optimistic to pessimistic situations. Besides taking different distributions of demands, we include different routing methods. Two strategies are developed, one using predetermined paths, and the other using dynamic routing. The deterministic strategy corresponds to a worse situation, where drivers have no global information about the traffic, for example due to the failure of the cellular network, and they enforce a decision at the beginning of the simulation ignoring any external dynamic information. Since the residents already have good knowledge about their cities, the path is arranged for each driver using the shortest path to the closest exit road. The second strategy corresponds to a better informed situation, where the drivers have some global traffic information with the update at a given frequency, and let the vehicles adjust to choose the fastest path rather than the shortest path. As present in Figure \ref{fig:maps}, both Paradise and Mill Valley have tree-like or forest-like maps, and there are not a lot of paths from the source to the destination. Therefore, the dynamic routing plays a more important role in balancing traffic loads on different exits rather than choosing different paths to the same exit road. We assume the local residents are familiar with the positions of the exits, so the success of the evacuation is not heavily depend on the dynamic routing. However, if a city road network is more grid-like, dynamic routing becomes more important to reduce gridlocks and lane merging. We discuss routing management for grid-like roads in section \ref{sec:discussion}. SUMO's automatic routing module provides such an ability to run the simulation with dynamic routing \cite{sumoautorouting}. The average speed of all cars passing through a given road is calculated periodically on the order of several minutes so that the average traveling time of that road is acquired. Each vehicle calculates the fastest path to the destination road or destination TAZ every minutes using Dijkstra's algorithm \cite{sumoautorouting}. \yuchen{ The weights of the edges in the algorithm are defined by the average travel time of the last five updates for the corresponding road segments.}
In table \ref{tab:scenarios}, column ``Routing" indicates the routing methods for different scenarios.

\subsection{Traffic management policy} \label{subsec:traffic_management_policy}

\begin{figure}[H]
\centering
\begin{subfigure}[t]{0.5\textwidth}
    \centering
    \includegraphics[height=6cm]{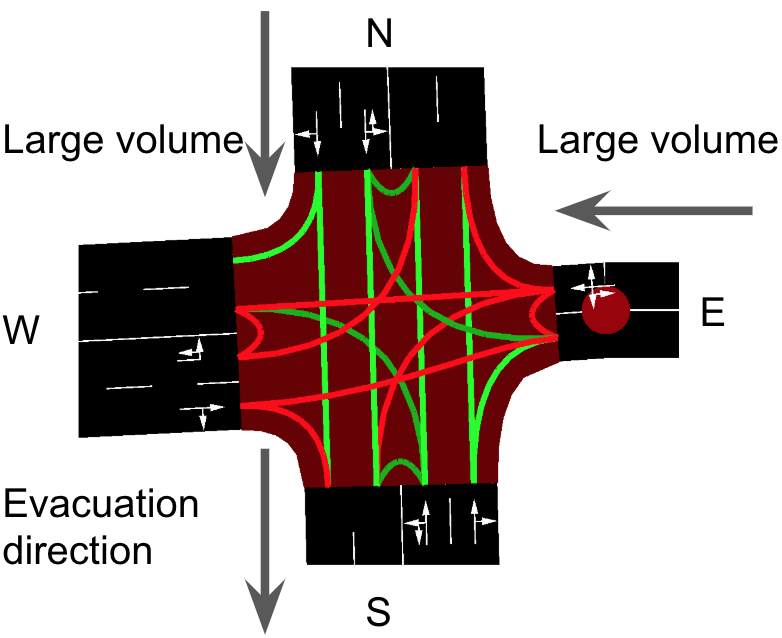}
    \caption{Normal traffic policy.}
\end{subfigure}%
\begin{subfigure}[t]{0.5\textwidth}
    \centering
    \includegraphics[height=6cm]{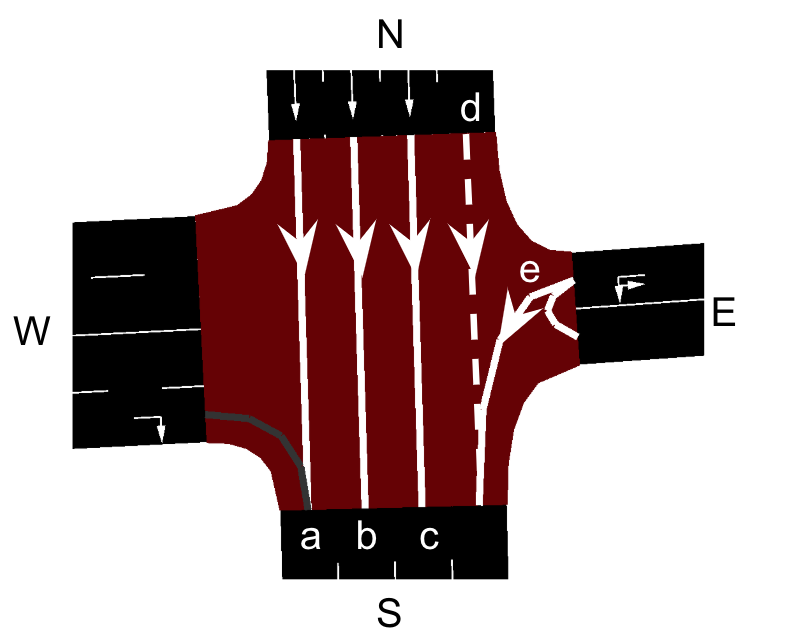}
    \caption{Proposed evacuation traffic policy.}
\end{subfigure}%

\begin{subfigure}[t]{0.5\textwidth}
    \centering
    \includegraphics[height=5.5cm]{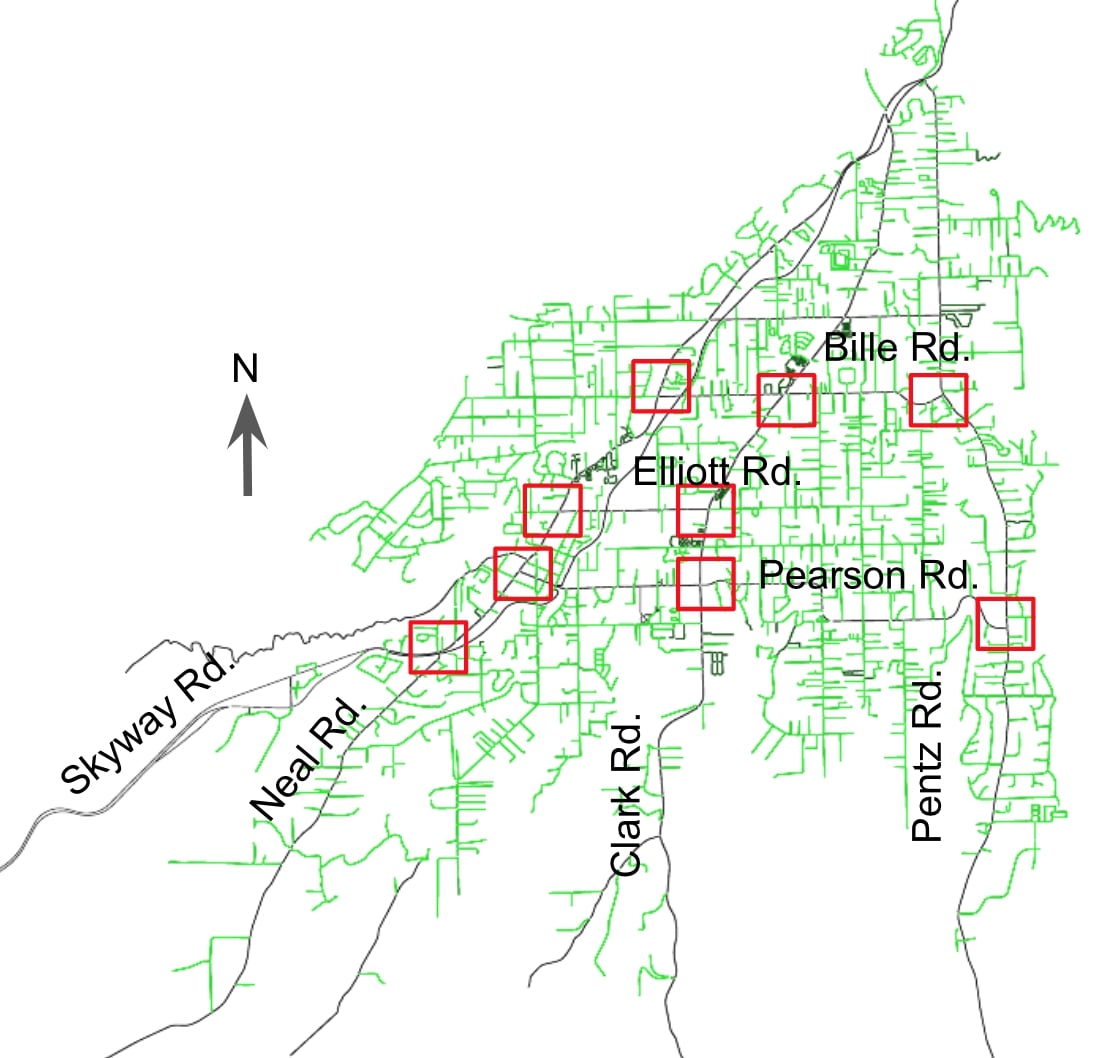}
    \caption{Changes of policy in Paradise}
\end{subfigure}%
\begin{subfigure}[t]{0.5\textwidth}
    \centering
    \includegraphics[height=5.5cm]{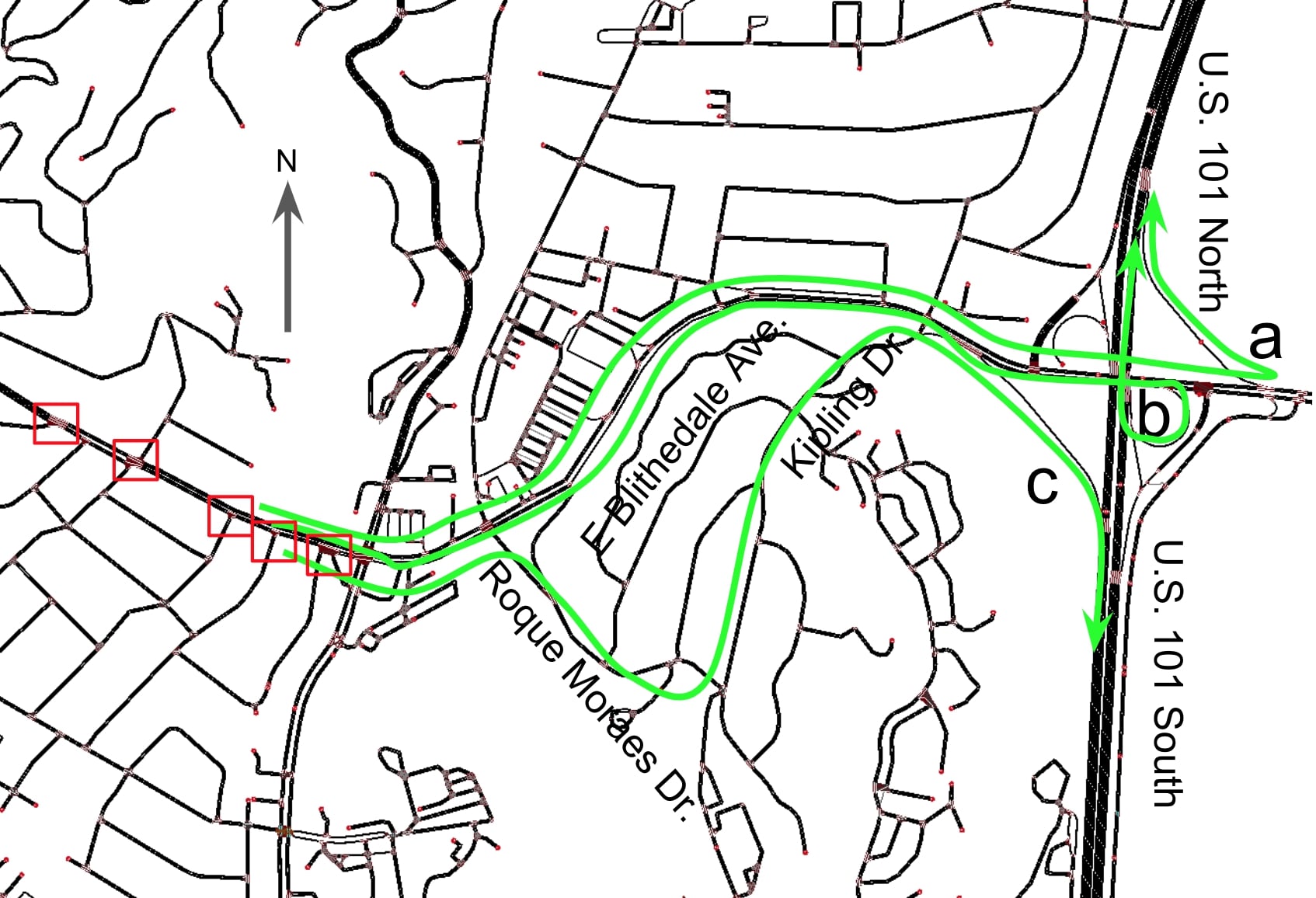}
    \caption{Changes of policy in Mill Valley}
\end{subfigure}%

\caption{Evacuation traffic management policies. (a) Normal traffic policy in a crossroad controlled by a traffic light. The colors represent the phases of the cross road. Red and green denote opposite conditions of the lanes at one phase, either green lanes pass and red lanes stop or the reverse. During the evacuation, there are large traffic volumes from the north side to the south side and from the east side to the west side. (b) The contraflow policy of the crossroad. Lane $a$ and $b$ remain the same. The directions of lane $c$ and $d$ are reversed. Lane $e$ comes from a residential road. There are two ways to handle lane merging. If lane $e$ carries a large traffic volume, lane $d$ at the intersection will be blocked, so that $e$ has reserved room for incoming cars. If lane $e$ has light traffic, lane $d$ will not be blocked at the intersection, and the cars from lane $e$ have to yield to the cars in lane $d$. (c) Changes to the crossroads in the evacuation policy in Paradise are labeled in red boxes. (d) Changes of the evacuation policy in Mill Valley. The mini map ranges from 4.5 to 6 km along the x-axis and from 3.5 km to 4.5 km along the y-axis relative to the map in Figure \ref{fig:maps} (b). Path $a$ represents the reverse road of E Blithedale Ave. Path $b$ represents the outgoing path of E Blithedale Ave. Path $c$ represents the outgoing path on E Blithedale Ave. but detoured through local road Roque Moraes Dr. These three paths enlarge the capacity relative to the normal policy. Lane changing is forbidden on those three paths so that vehicles can speed up and avoid collisions. Red boxes mark the crossroads modified by the contraflow method.}
\label{fig:new_traffic_policies}
\end{figure}

The proposed evacuation traffic management policies are motivated by the issues found in the simulation speed map and simulation animation using \textit{sumo-gui}. See Figure \ref{fig:speed_maps_paradise} and \ref{fig:speed_maps_millvalley} for details, which are further elaborated upon in section \ref{sec:results} \textit{speed maps}. The bottlenecks are usually related to the crossroads along the arterial roads, as traffic lights, lane merging, and stop signs at the crossroads can all slow down the traffic flow. The normal traffic rules are not efficient because of the stop-and-go traffic control settings. The aggressive driving behaviors under emergency situations can further lead to gridlock or more collisions \cite{xie2011lane}. Subsequently, the blocked vehicles quickly form traffic jams that propagate to upstream areas. Those intersections are designed for daily usage to handle traffic flow from many directions, and are not suitable for high volume traffic flow in a one direction during evacuations. Another factor that influences traffic flow is lane merging and changing. This behavior happens when a vehicle is trying to get into a higher priority road from a lower priority road, or is trying to change its path using different lanes. During evacuations with large traffic flows, both lane merging and lane changing can be very difficult as there is no extra space in the target lane. These lead us to design the evacuation traffic management policies to fix the crossroads problems.

\yuchen{We seek simple and consistent traffic management policies to fix the above issues, which require as little supervision as possible, so that they are less sensitive to power outages, driver behaviors, police supervision, public education, etc. Also, they should ideally not require additional construction and be implementable using existing infrastructure in order to be more practical, economical, and able to be deployed immediately. We focus on changing the traffic rules along the arterial roads, and leaving the local roads' rules as they are. This approach requires less supervision for local communities and takes advantage of the potential capacities of the arterial roads. In addition, it makes evacuation policy design easier and reduces the search space of tuning options.} Our new design of traffic management policies is demonstrated in Figure \ref{fig:new_traffic_policies}.

The strategy for the evacuation traffic management policy is demonstrated in Figure \ref{fig:new_traffic_policies}. Figure \ref{fig:new_traffic_policies} (a) shows the evacuation traffic flow under normal traffic control. During the evacuation, most vehicles will go outbound in one direction. Lane switching is controlled by the traffic light. The phases of the traffic light are indicated by the colors of the roads. Green and red represent opposite lane statuses, either green lanes pass and red lanes stop or the reverse. If there is a large volume of traffic flow from $N$ to $S$ and no traffic from $S$ to $N$, then the traffic light does not need to spare time to let the vehicles pass from $W$ to $E$, from $E$ to $W$ or from $S$ to $N$. A natural idea to improve the evacuation efficiency is removing unnecessary connections so that the traffic control only serves the large volume traffic directions. Figure \ref{fig:new_traffic_policies} (b) shows the evacuation policy. The white lanes are the modified lane connections. Lane $c$ and $d$ used to connect from $S$ to $N$ are reversed. Lane $c$ and $d$ are reversed to carry more traffic from the top side to the bottom side. Reversing the inbound lanes can significantly increase or even double the roads' capacity. Vehicles from lane $e$ have to yield to vehicles on lane $d$. When the traffic flow from lane $e$ is light, this strategy works well. However, lane $d$ may still block the traffic from lane $e$ as it has higher priority. Another way of the modification is similar, except for the changes of lane $d$ and lane $e$. To handle large incoming traffic volumes from $E$ to $S$ and to avoid conflicts in lane merging, lane $d$ at the crossroad is blocked (not blocked in the upstream road) in order to supply more room for lane $e$, so it is easier for the vehicles from lane $e$ to merge into the main stream. If there is a large volume traffic stream from $W$ to $S$, the modification is similar. The evacuation policy benefits from much less stopping, more lanes in one direction, and less lane merging conflicts. These types of network reconfiguration strategies are known as \textit{contraflow}, and show impressive performance in previous work \cite{wolshon2001one,shekhar2006contraflow,tuydes2006tabu,meng2008microscopic,xie2011lane,pyakurel2015models}. We only change the policies on the arterial roads to make the evacuation policy as simple as possible. The residential roads remain bidirectional as before to avoid confusion, and this choice can also help keep the network robust since if one arterial road is blocked, drivers can go back to take another exit.

For Paradise, the exits are the arterial roads Skyway Rd., Neal Rd., Clark Rd., and Pentz Rd. as shown in Figure \ref{fig:maps} (a) and Figure \ref{fig:new_traffic_policies} (c). We make the following changes for the evacuation traffic management policy. The modified crossroads are marked in Figure \ref{fig:new_traffic_policies} (b).
\begin{enumerate}
\item All the inbound arterial roads are reversed except for Neal Rd., which is saved for emergency and rescue vehicles. 
\item The intersections along Skyway Rd. are modified using the contraflow method. And the eastern-most lanes at the crossroads are blocked to admit the traffic flow from Bille Rd., Elliott Rd. and Pearson Rd.
\item The intersections along Clark Rd. are modified using the contraflow method, and the eastern-most lanes at the crossroads are blocked to admit the traffic flow from Bille Rd., Elliott Rd., and Pearson Rd.
\item The intersections along Pentz Rd. are modified using the contraflow method, and the western-most lanes at the crossroads are blocked to admit the traffic flow from Bille Rd. and Pearson Rd.
\end{enumerate}

We make change 1 to double the number of lanes for outbound direction. Change 2 is to let Skyway Rd. better deal with the large incoming traffic from the east and north. Change 4 is for Pentz Rd. to better handle the heavy traffic from the west and the north. In change 3, the eastern-most lanes at the crossroads along Clark Rd. are blocked to handle large incoming traffic streams because the residential area between Clark Rd. and Pentz Rd. is larger than that between Clark Rd. and Skyway Rd.. Pearson Rd., Elliott Rd., and Bille Rd. are the only ways to get out, so they carry heavier traffic than the roads on the west side of the Clark Rd. If both the western-most lanes and eastern-most lanes are blocked alone the Clark Rd., they will bottleneck traffic, as four lanes merge into two lanes at the crossroads. The traffic between Clark Rd. and Skyway Rd. is encouraged to take Skyway Rd. out, which is enlarged to four lanes after modification. So we choose only to block the east lanes instead of the west lanes at the intersections along the Clark Rd. A comparison between before and after the modification of the traffic management policy is presented in scenarios 1, 2, and 3 in Table \ref{tab:scenarios}. The speed maps and evacuation performance comparison between scenarios 1, 2, and 3 are shown in section \ref{sec:results}. Note that residential roads are not changed to keep the policy robust in the case that drivers need to turn back to other exits. Also, they are not bottlenecks when compared with arterial roads (see Figures \ref{fig:speed_maps_paradise} and \ref{fig:speed_maps_millvalley}). As for whether to reserve one lane for incoming cars from the residential roads, we tune the policy manually according to empirical observations in \textit{sumo-gui} and the speed map. There are potential optimal methods to tune the contraflow analytically \cite{xie2011lane}, or via empirical trial-and-error \cite{kheterpal2018flow}. However, they require information about the traffic flow and satisfying certain assumptions, which are either unknown before the simulation or too complicated for our case, given that the scenario includes only a few intersections.

\retouch{For Mill Valley, the exits are the arterial roads E Blithedale Ave. and Miller Ave in the central area and Shoreline Hwy. in the south side of the city. Accounting for the management of and alternatives to E Blithedale Ave. is important, as the road is a principal exit for a large part of the city. The evacuation traffic management policies are shown in Figure \ref{fig:new_traffic_policies}. We make the following changes for the evacuation traffic management policy.}
\begin{enumerate}
\item The inbound E Blithedale Ave. is reversed. Inbound road of Shoreline Hwy. is not changed as the rescue entrance.
\item The reversed path $a$ of E Blithedale Ave. in Figure \ref{fig:new_traffic_policies} (d) takes the U.S. 101 N with the U-turn ramp as the exit. This path mainly takes the traffic from the north side of the city.
\item The path $b$ of E Blithedale Ave. in Figure \ref{fig:new_traffic_policies} (d) takes the U.S. 101 N as the exit with the circular ramp. This path mainly takes the traffic beyond the downtown area.
\item The path $c$ of E Blithedale Ave. in Figure \ref{fig:new_traffic_policies} (d) is detoured to Roque Moraes Dr. and the exits at Kipling Dr. near the entrance to the U.S. 101 S.
\item All traffic lights are removed along E Blithedale Ave. The contraflow method is applied at several crossroads marked by the red boxes in Figure \ref{fig:new_traffic_policies} (d).
\item Lane changing is forbidden on E Blithedale Ave. on the section shown in Figure \ref{fig:new_traffic_policies} (d).
\end{enumerate}

We make change 1 to double the number of lanes for the outbound direction. As path $b$ has only one lane in the middle section of the E Blithedale Ave., \retouch{ the simulation indicates that it} slows the efflux traffic flow due to lane merging, and there is no further room for improvement. Path $c$ is assigned as complementary to path $b$. Part of the the outgoing traffic flow on E Blithedale Ave. is detoured to Roque Moraes Dr., and takes U.S. 101 S as the exit. The contraflow method is applied at several crossroads marked by the red boxes. The southern-most lanes are reserved for the traffic from the downtown area. Path $a$ mainly takes the traffic from the north side of the town. Path $b$ is arranged to takes the traffic from the part west to the downtown area. Path $c$ carries the traffic from the downtown area. Traffic in the southern part of the city will take Shoreline Hwy. The contraflow made in change 5 guarantees that the vehicles can quickly merge into E Blithedale Ave. In changes 2, 3, and 4, paths $a, b, c$ take separate lanes and separate exits, so they reduce the lane changing and merging as much as possible to let drivers travel more quickly. For change 6, to avoid potential collisions or other accidents, lane changing is forbidden on the roads marked by the green arrow in Figure \ref{fig:new_traffic_policies} (d). In the southern part of Mill Valley, Shoreline Hwy. is the only way out and only way in, and it has one lane for both directions. Therefore, the traffic management policy is not modified at that location. The inbound road of Shoreline Hwy. is not reversed to keep it as the emergency vehicle entrance. Shoreline Hwy. covers a smaller residential area and carries less traffic than E Blithedale Ave., so that is the reason why it is chosen as the rescue entrance. Scenarios 8, 9, and 10 in Table \ref{tab:scenarios} compare the evacuation efficiency between the evacuation policy with the baseline scenarios. The speed maps and evacuation performance comparison between scenarios 8, 9 and 10 are shown in section \ref{sec:results}. Similar to the case study of Paradise, we manually design and test the evacuation policy since the network layout of Mill Valley is simple.

\subsection{Robustness of the traffic management policy} \label{subsec:scenarios_robustness}
One of the the most important lesson from the Camp Fire in Paradise is that we should never underestimate the immense potential of a small fire and its rapidly spreading speed. The fire already reached the town two hours after starting seven miles away, while the full-scale evacuation had not been announced yet \cite{paradise2018firereachtime,paradise2018timeline}. This weather-accelerated behavior brings challenges to evacuation planning. For example, exits can be closed by surrounding fires, fallen trees, abandoned vehicles, and other obstacles. It is important to take these risks into account for traffic management policy planning. In our simulation scenarios, we study the consequences of blocking the main exit roads. For Paradise, Pentz Rd. is blocked at different times (see scenario 7 in Table \ref{tab:scenarios}).

\retouch{Unlike Paradise, which in principle has multiple exits, Mill Valley has one main egress for the northern part of the city, E Blithedale Ave., and so it is ideal to keep its traffic flowing. In the proposed traffic management policy, we enlarge the capacity of E Blithedale Ave. by arranging paths $a, b, c$ shown in Figure 
\ref{fig:new_traffic_policies} (d).} If one of the paths is blocked, the other two can act as backups. In scenario 14 in Table \ref{tab:scenarios}, the main road (path $b$ in Figure \ref{fig:new_traffic_policies} (d)) is blocked in order to test the robustness of the new design. \yuchen{ We evaluate the consequence of blocking a certain road using the evacuation efficiency. The result is shown in section \ref{sec:results} \textit{Robustness of the evacuation traffic management policy}. }

\section{Results} \label{sec:results}
\begin{figure}[H]
\centering
\begin{subfigure}[t]{0.5\textwidth}
    \centering
    \includegraphics[height=5cm]{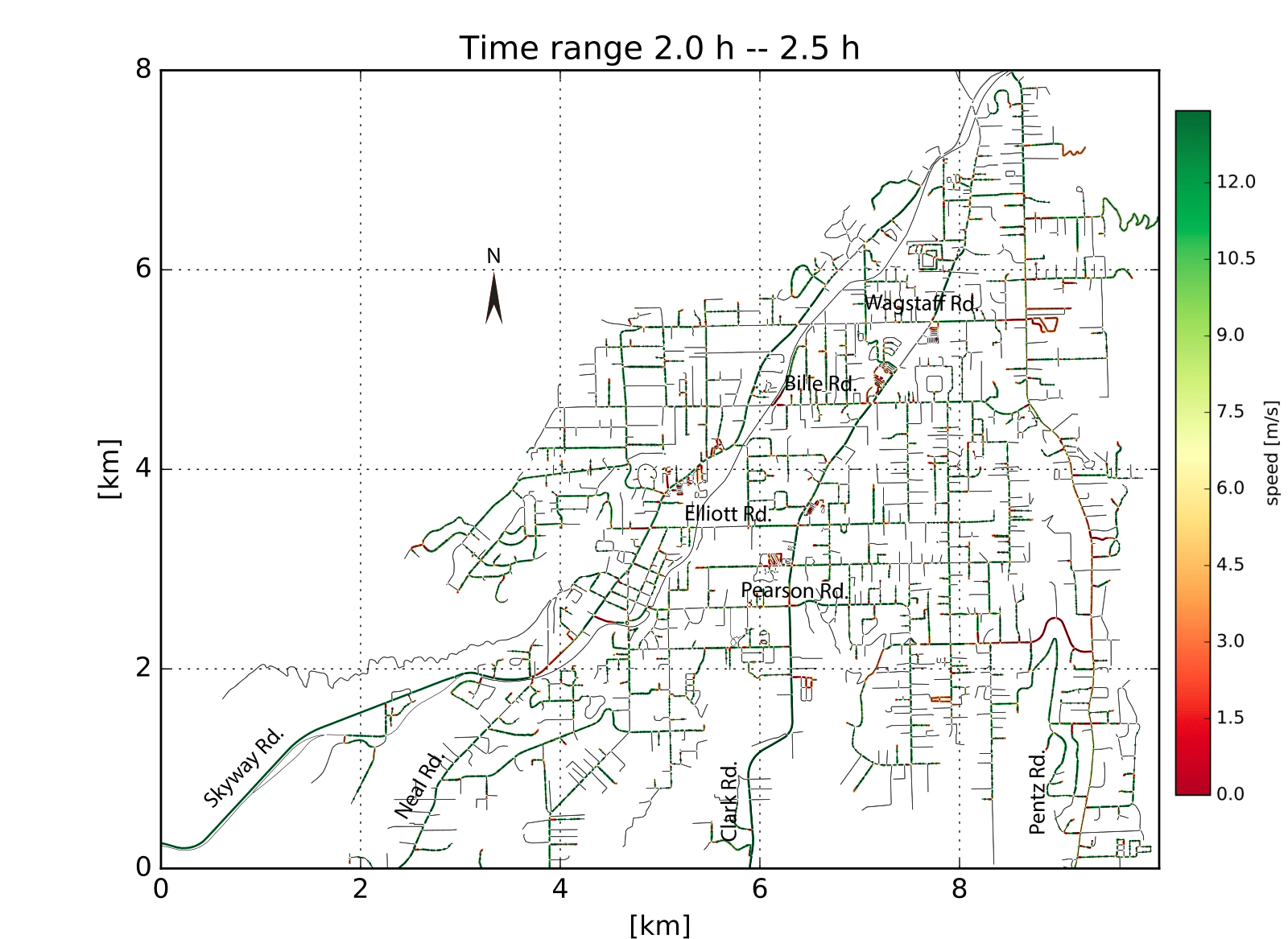}
    \caption{Scenario 1}
\end{subfigure}%
\begin{subfigure}[t]{0.5\textwidth}
    \centering
    \includegraphics[height=5cm]{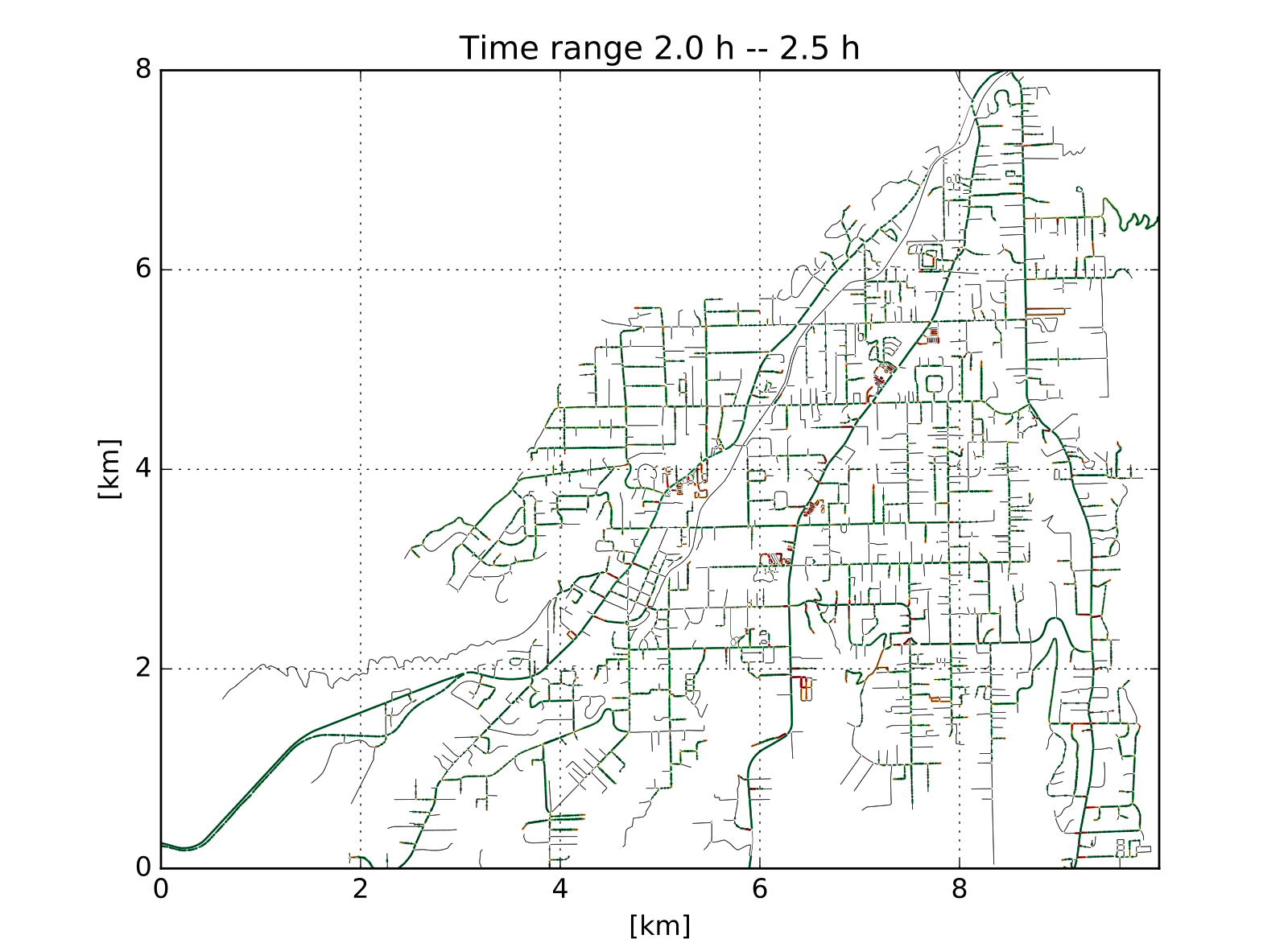}
    \caption{Scenario 3}
\end{subfigure}%

\begin{subfigure}[t]{0.5\textwidth}
    \centering
    \includegraphics[height=5cm]{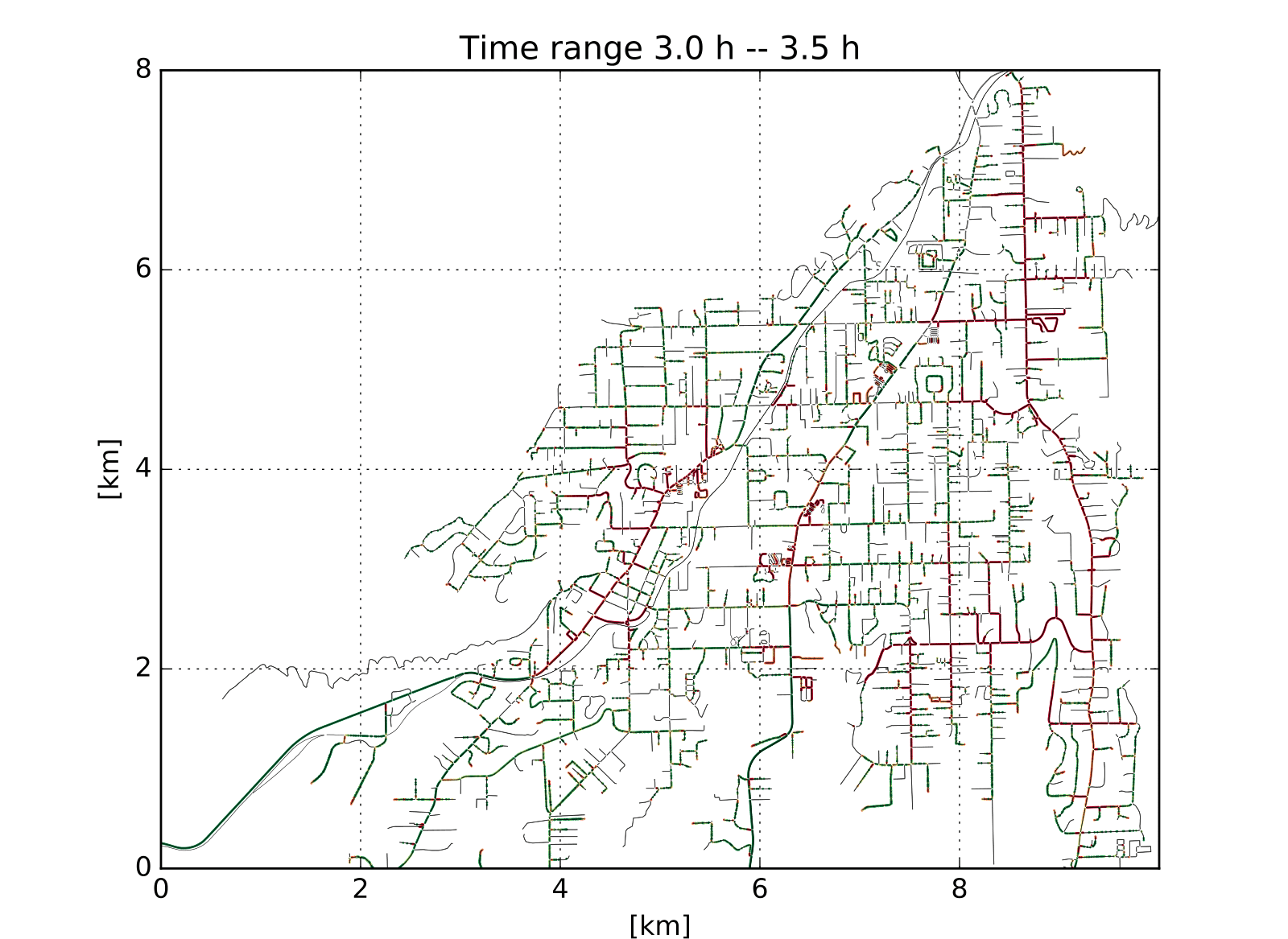}
    \caption{Scenario 1}
\end{subfigure}%
\begin{subfigure}[t]{0.5\textwidth}
    \centering
    \includegraphics[height=5cm]{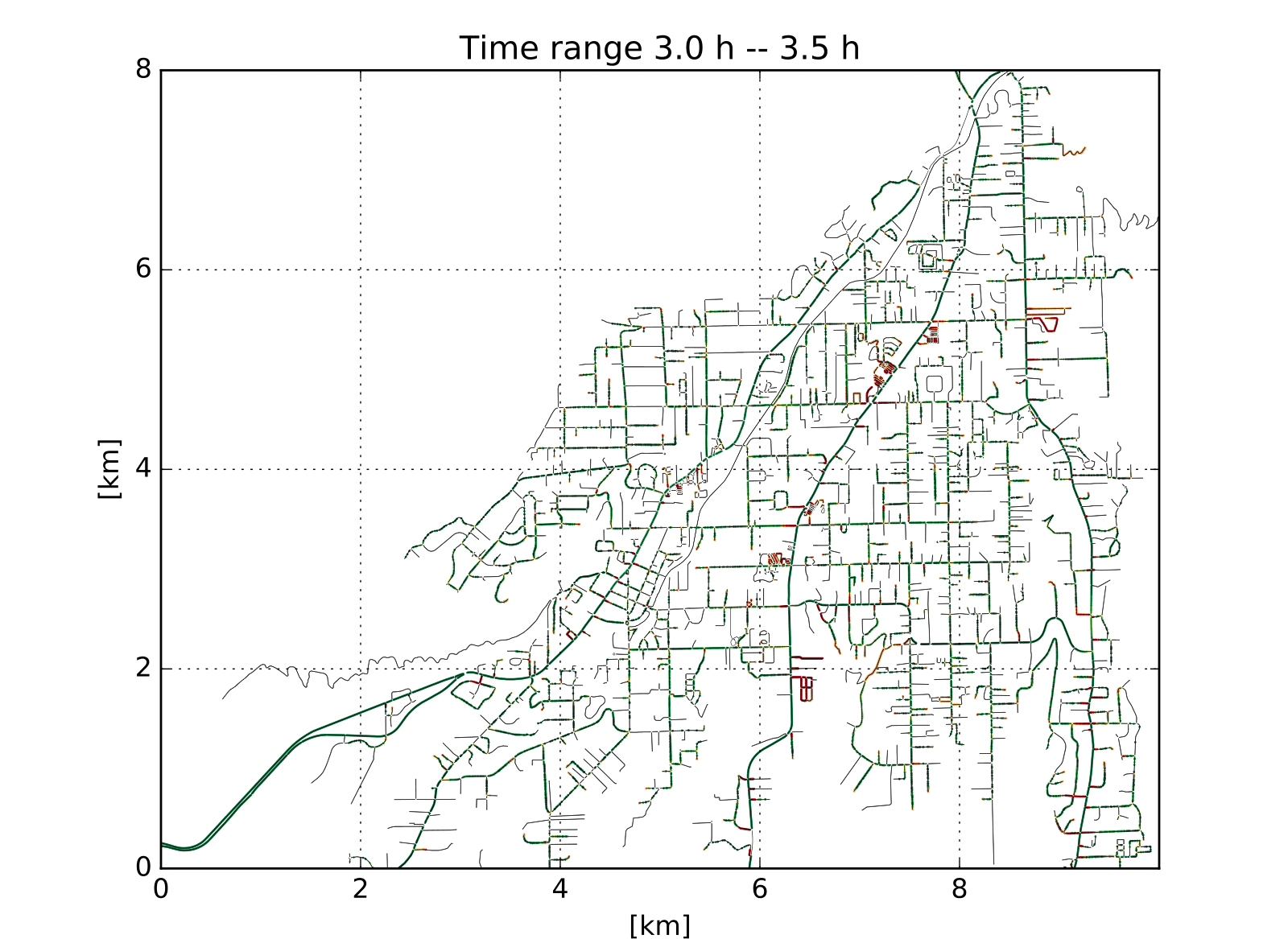}
    \caption{Scenario 3}
\end{subfigure}%

\begin{subfigure}[t]{0.5\textwidth}
    \centering
    \includegraphics[height=5cm]{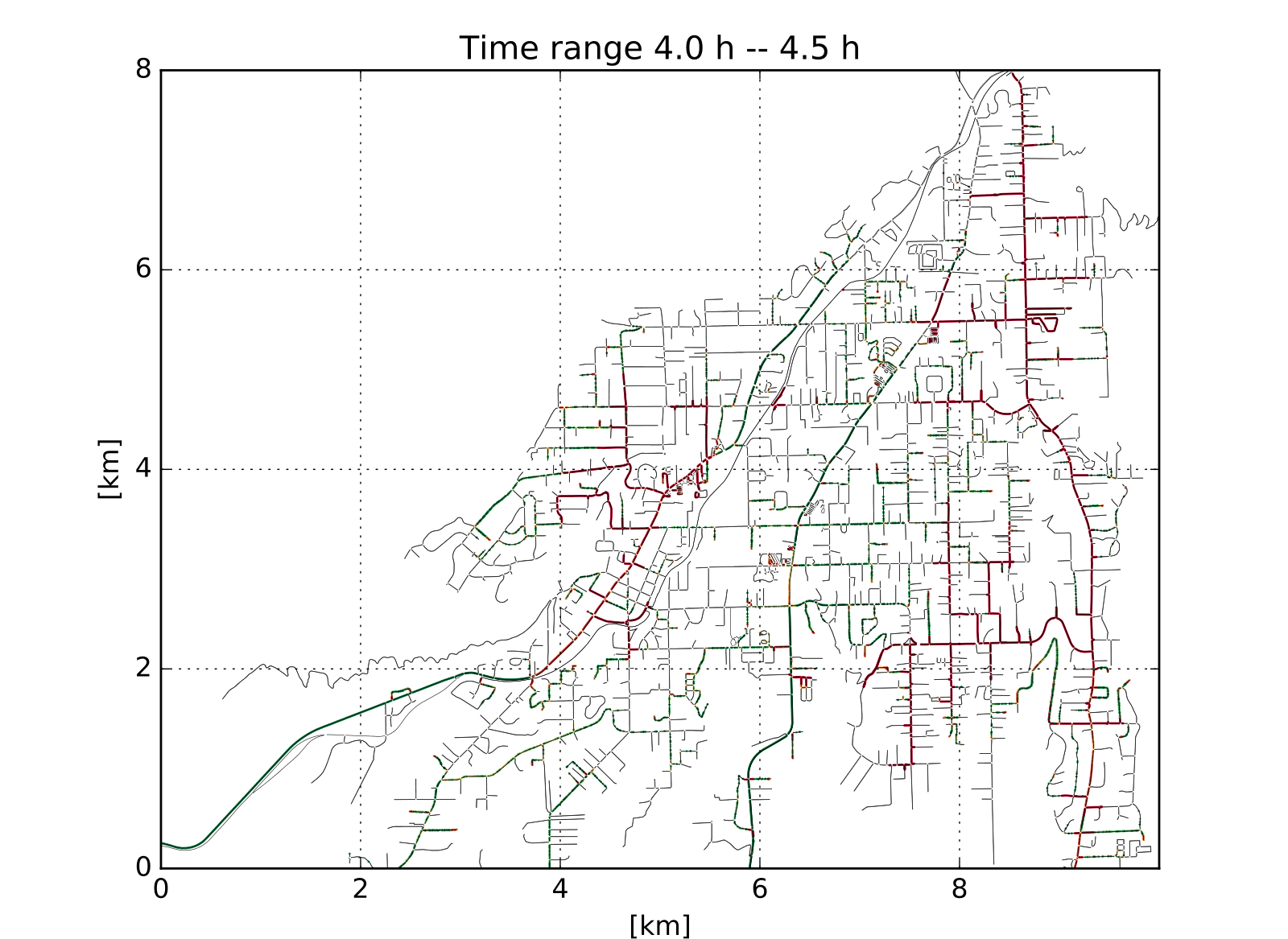}
    \caption{Scenario 1}
\end{subfigure}%
\begin{subfigure}[t]{0.5\textwidth}
    \centering
    \includegraphics[height=5cm]{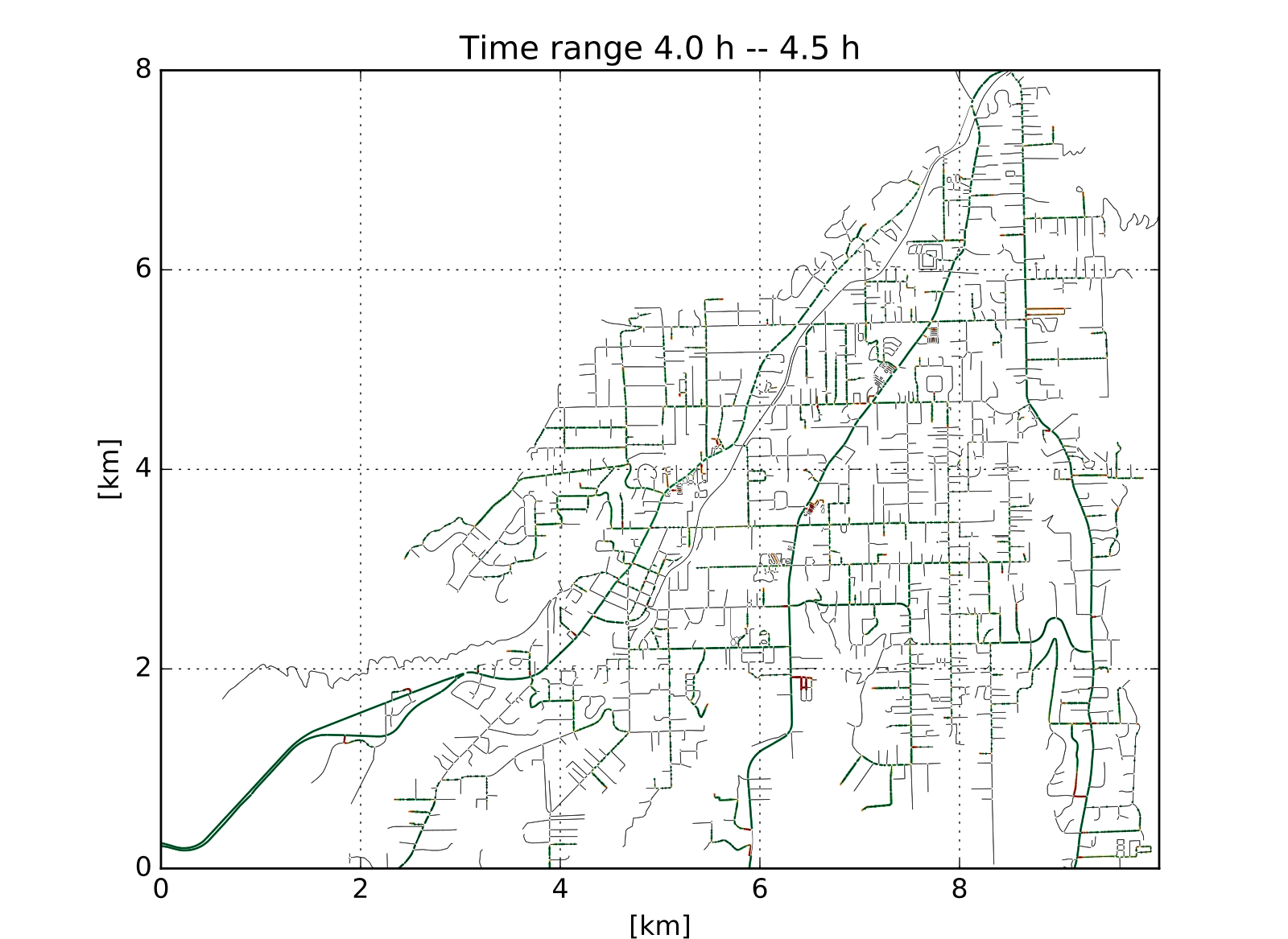}
    \caption{Scenario 3}
\end{subfigure}%

\begin{subfigure}[t]{0.5\textwidth}
    \centering
    \includegraphics[height=5cm]{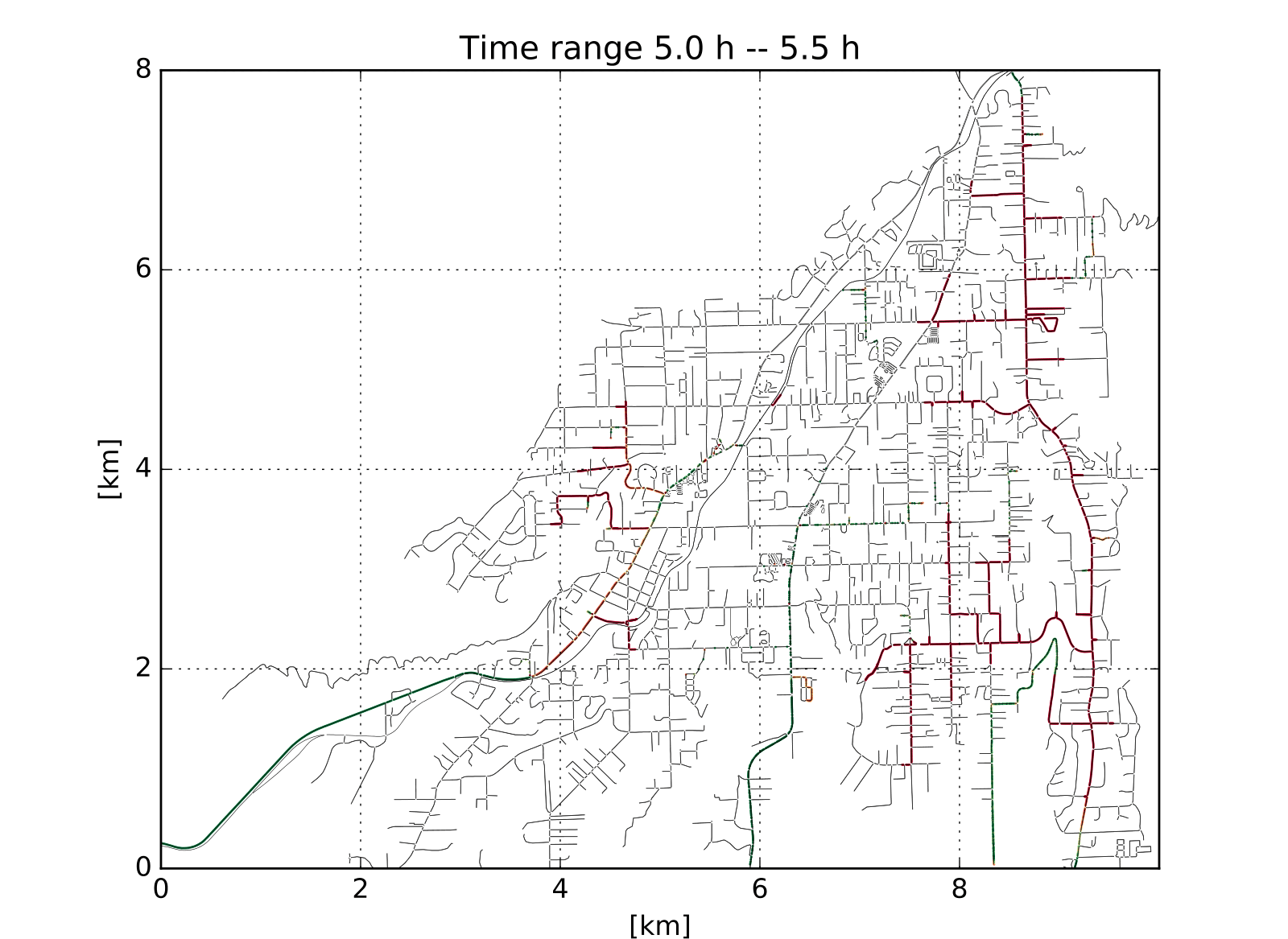}
    \caption{Scenario 1}
\end{subfigure}%
\begin{subfigure}[t]{0.5\textwidth}
    \centering
    \includegraphics[height=5cm]{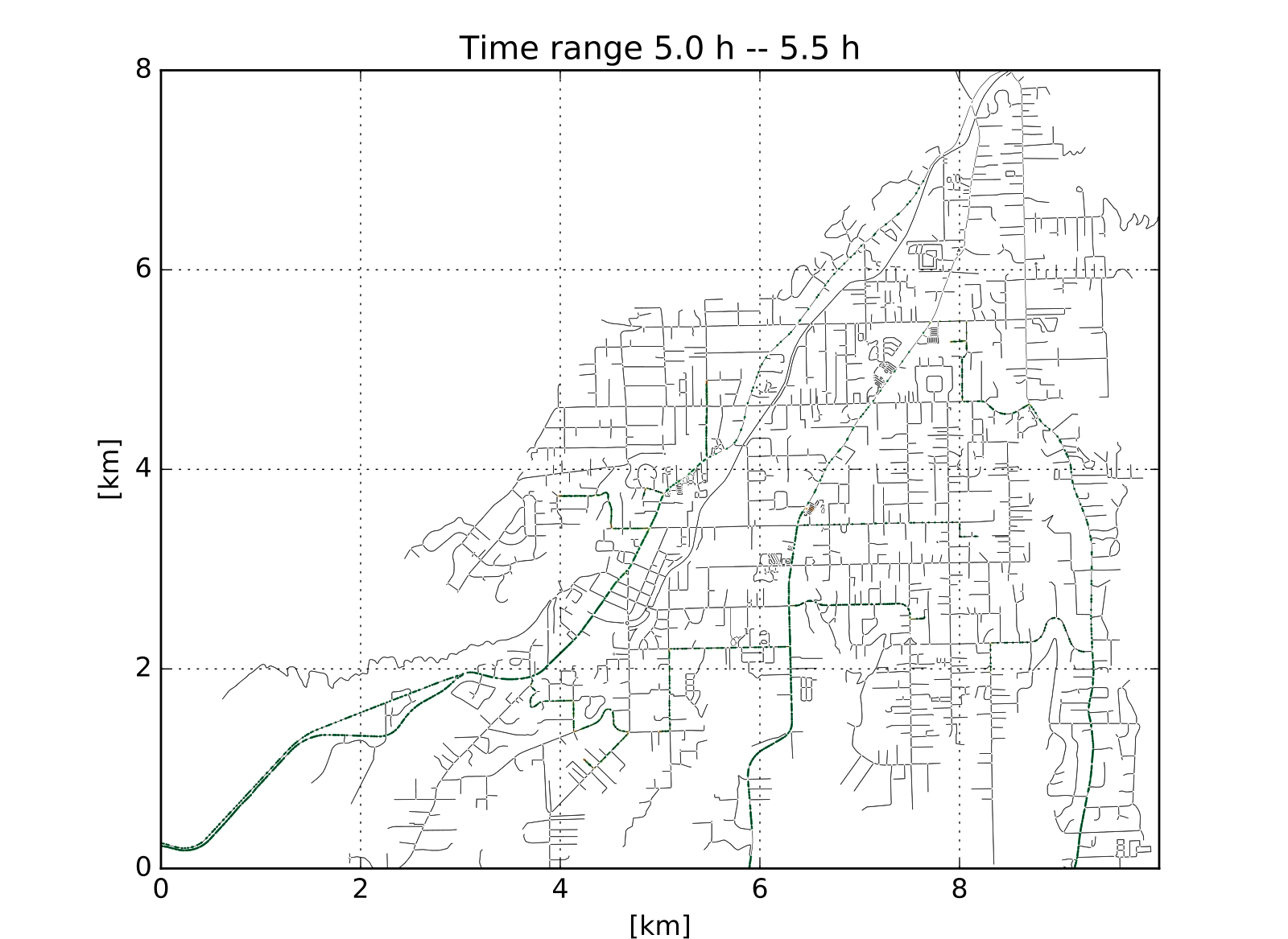}
    \caption{Scenario 3}
\end{subfigure}%

\caption{The speed maps for Paradise. The demands temporal distribution standard deviation is 0.7 h. (a, c, e, g) Scenario 1 in Table \ref{tab:scenarios} with shortest path routing and normal traffic management policy. (b, d, f, h) Scenario 3 with automatic routing and contraflow traffic management policy. }
\label{fig:speed_maps_paradise}
\end{figure}

\begin{figure}[H]
\centering

\begin{subfigure}[t]{0.5\textwidth}
    \centering
    \includegraphics[height=5cm]{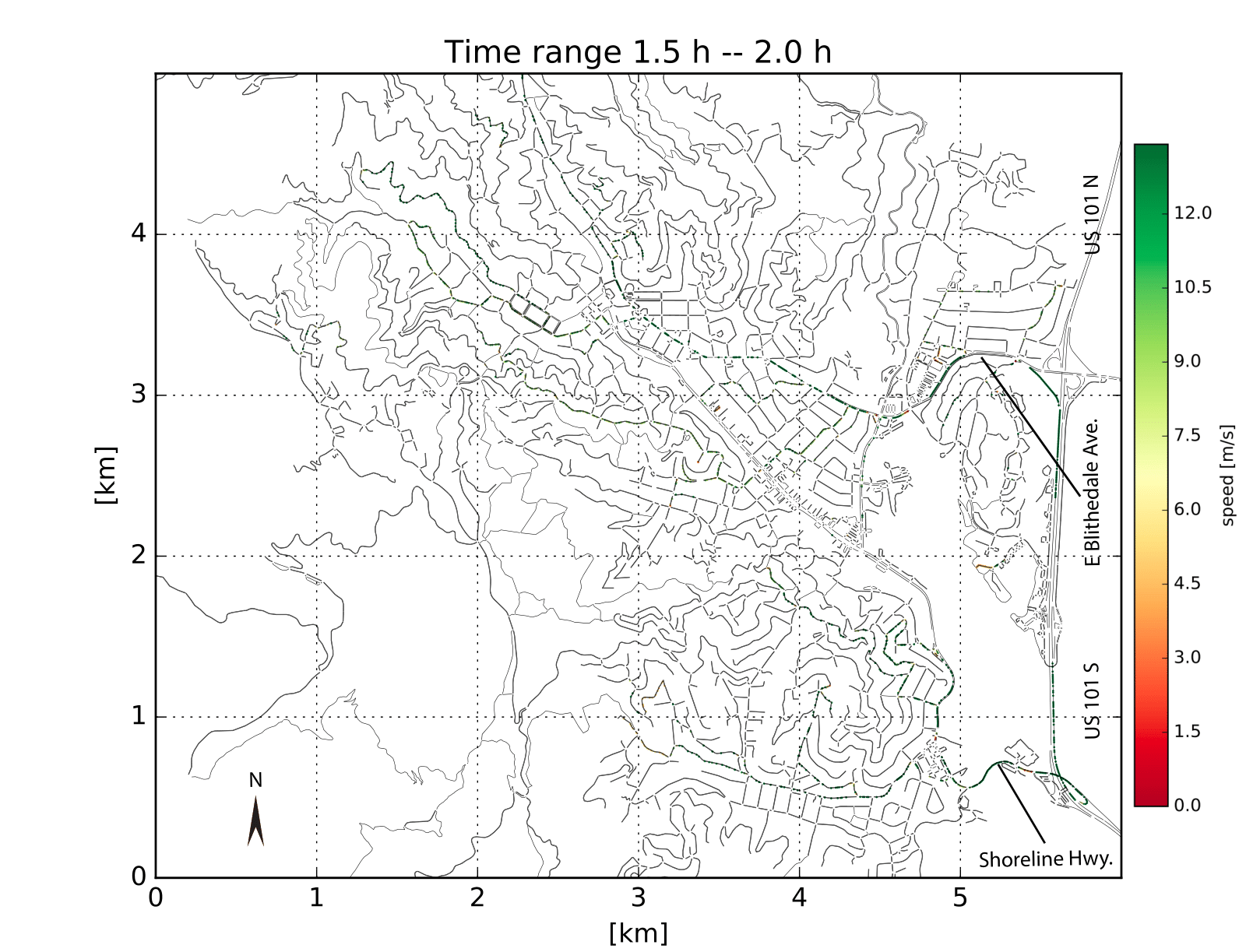}
    \caption{Scenario 8}
\end{subfigure}%
\begin{subfigure}[t]{0.5\textwidth}
    \centering
    \includegraphics[height=5cm]{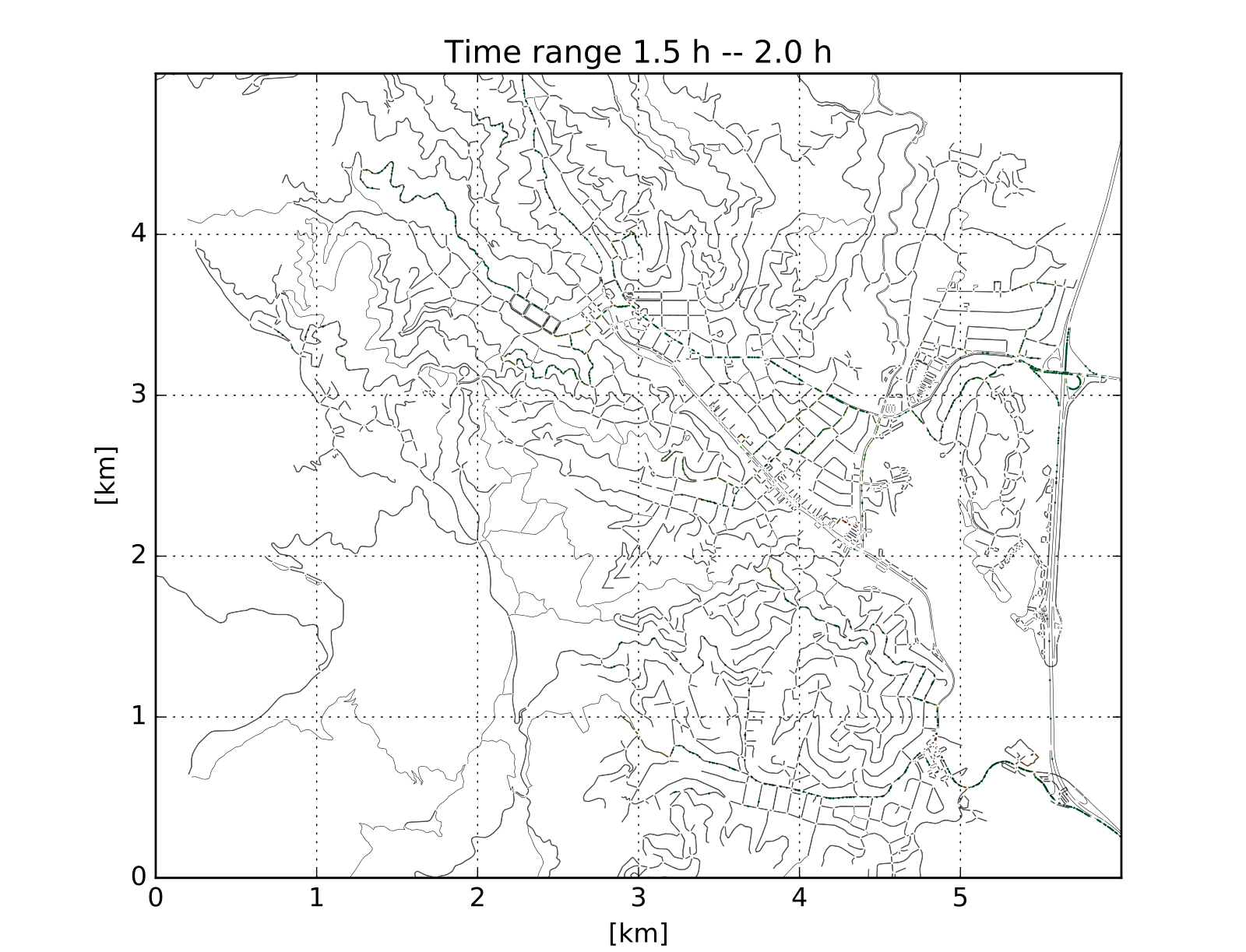}
    \caption{Scenario 10}
\end{subfigure}%

\begin{subfigure}[t]{0.5\textwidth}
    \centering
    \includegraphics[height=5cm]{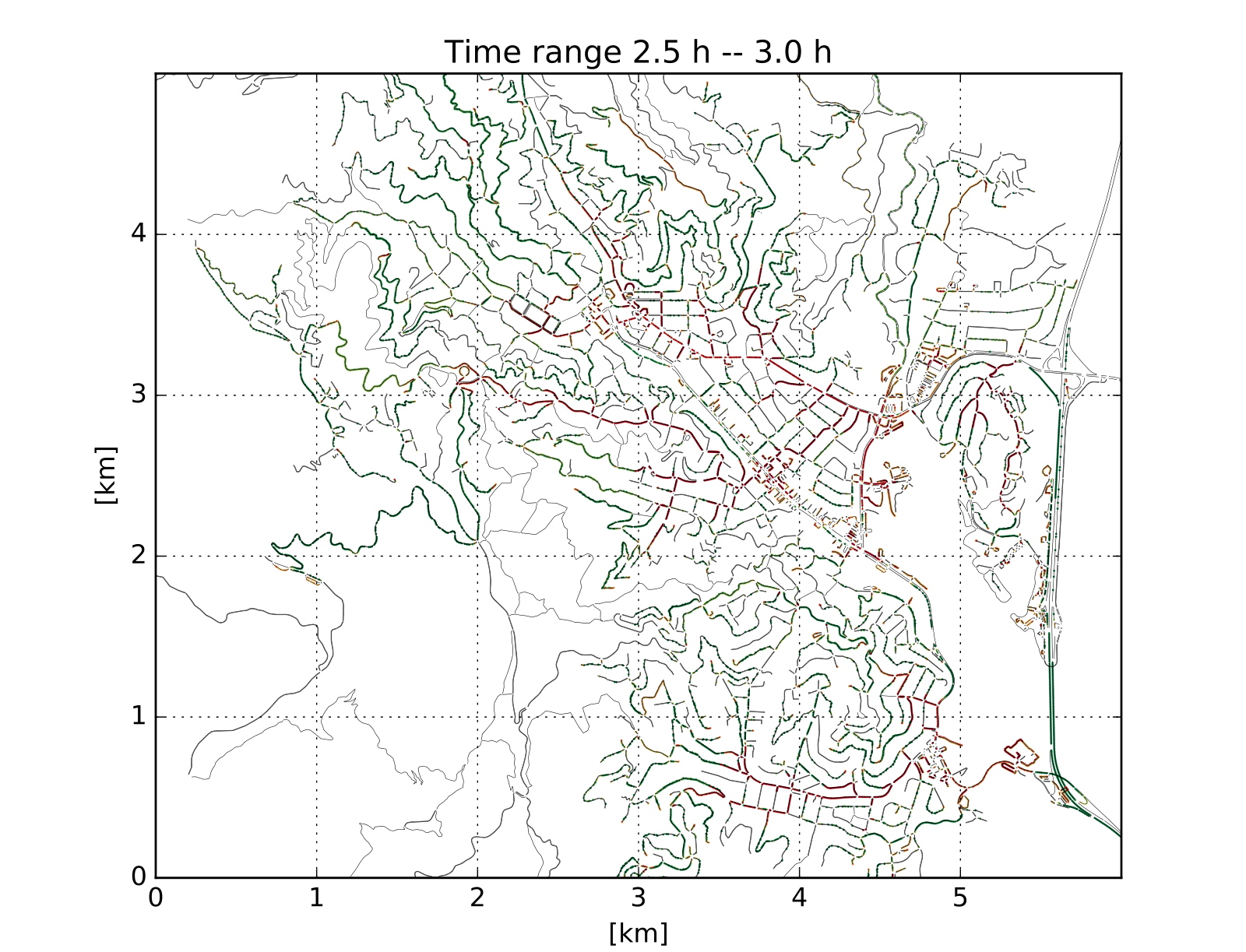}
    \caption{Scenario 8}
\end{subfigure}%
\begin{subfigure}[t]{0.5\textwidth}
    \centering
    \includegraphics[height=5cm]{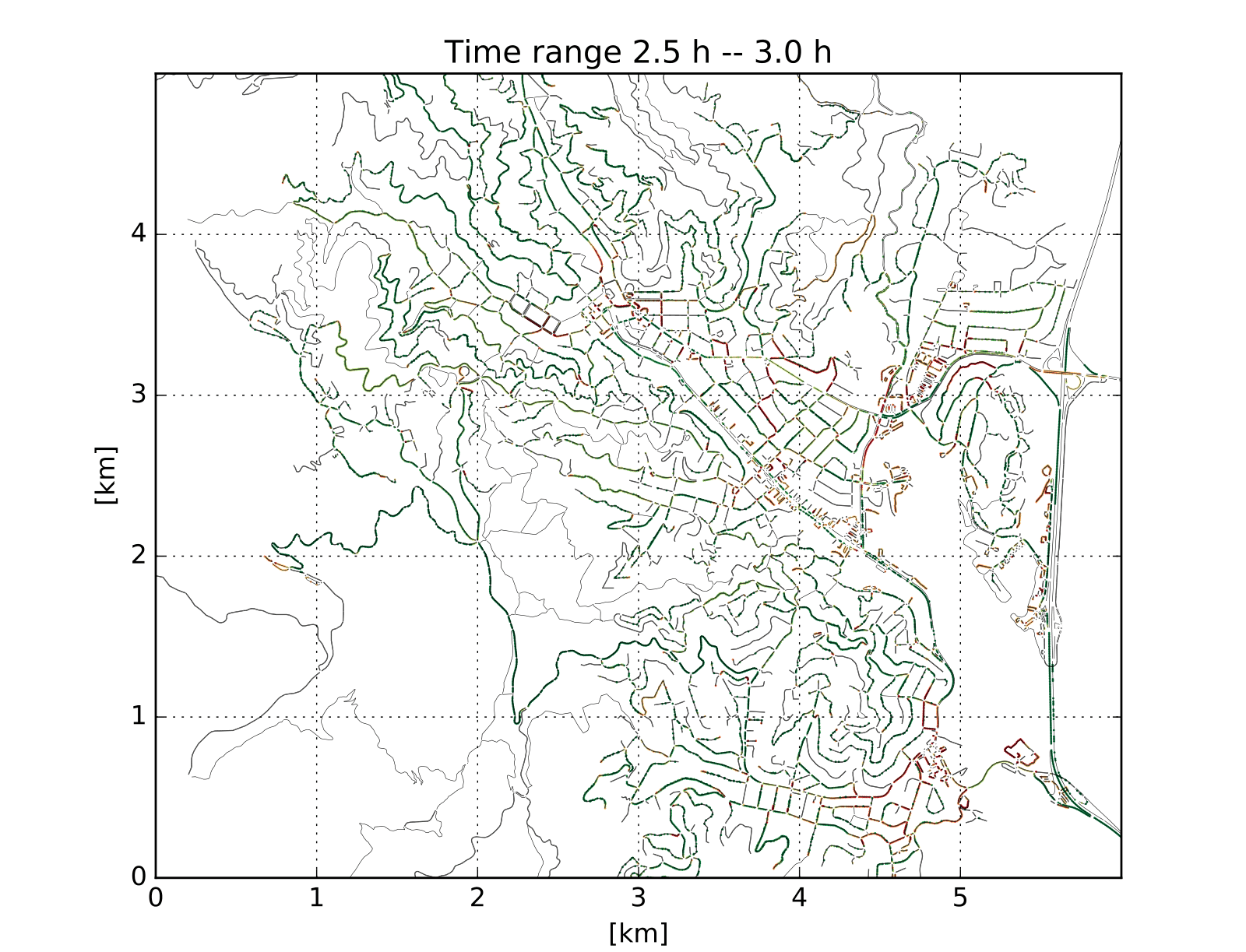}
    \caption{Scenario 10}
\end{subfigure}%

\begin{subfigure}[t]{0.5\textwidth}
    \centering
    \includegraphics[height=5cm]{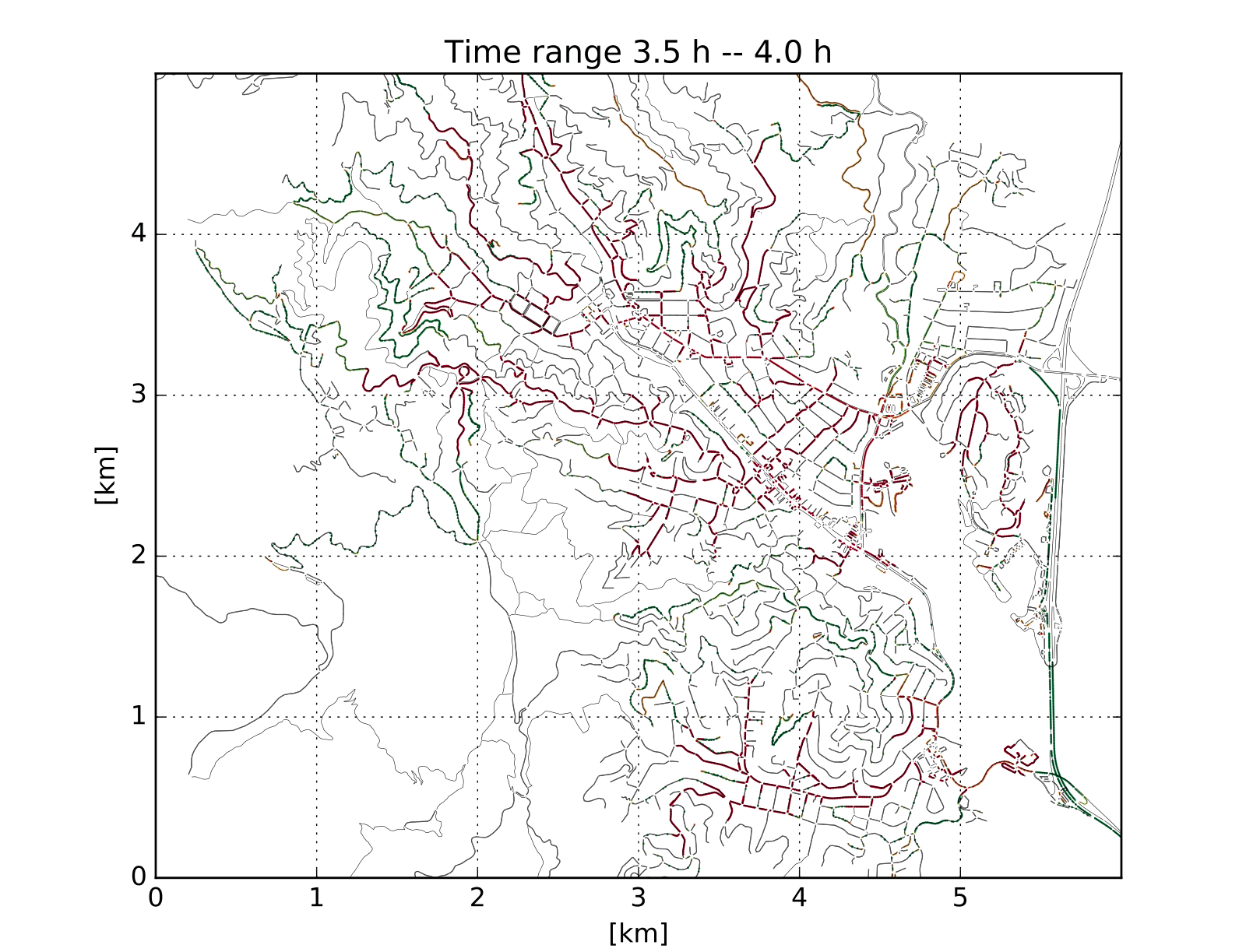}
    \caption{Scenario 8}
\end{subfigure}%
\begin{subfigure}[t]{0.5\textwidth}
    \centering
    \includegraphics[height=5cm]{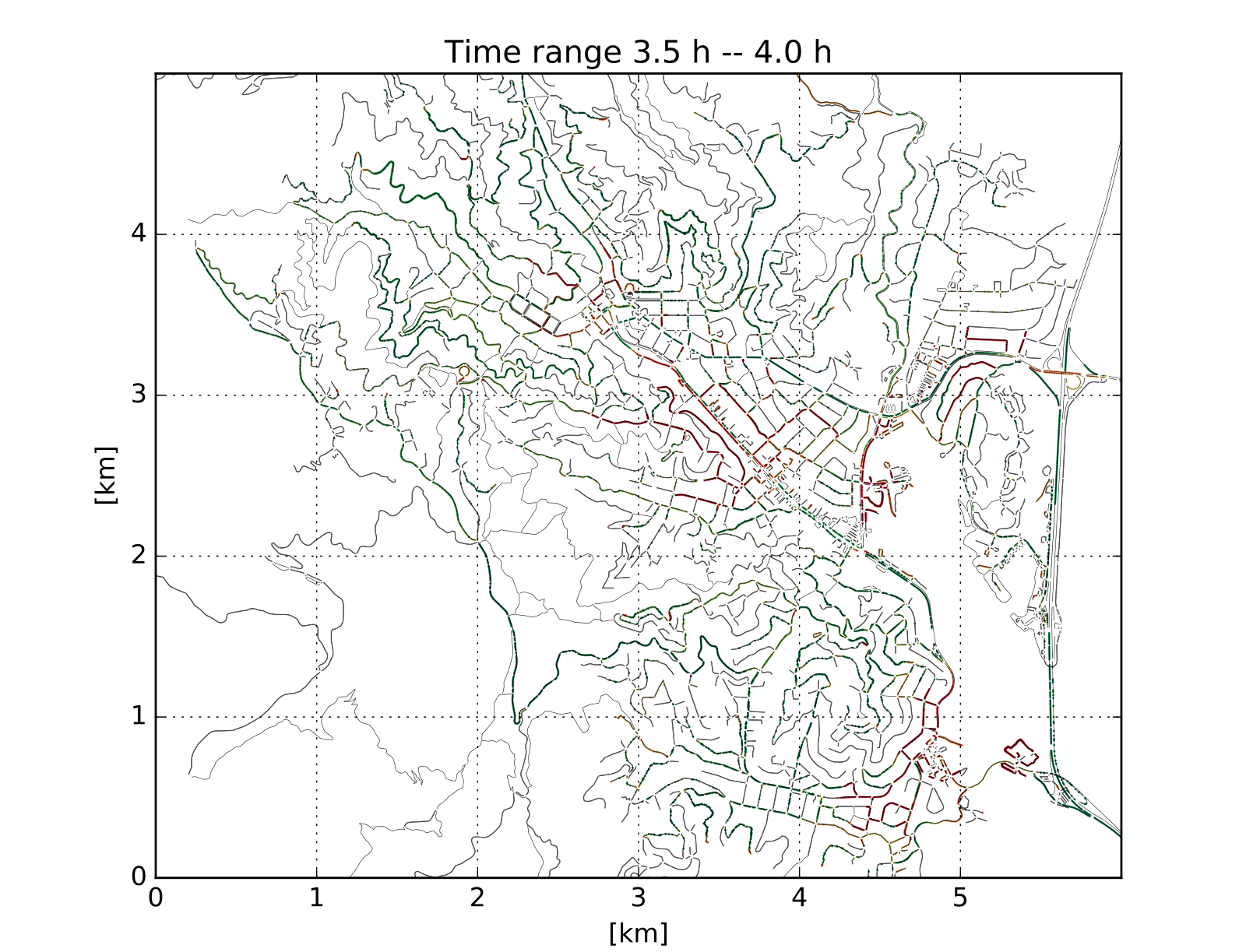}
    \caption{Scenario 10}
\end{subfigure}%

\begin{subfigure}[t]{0.5\textwidth}
    \centering
    \includegraphics[height=5cm]{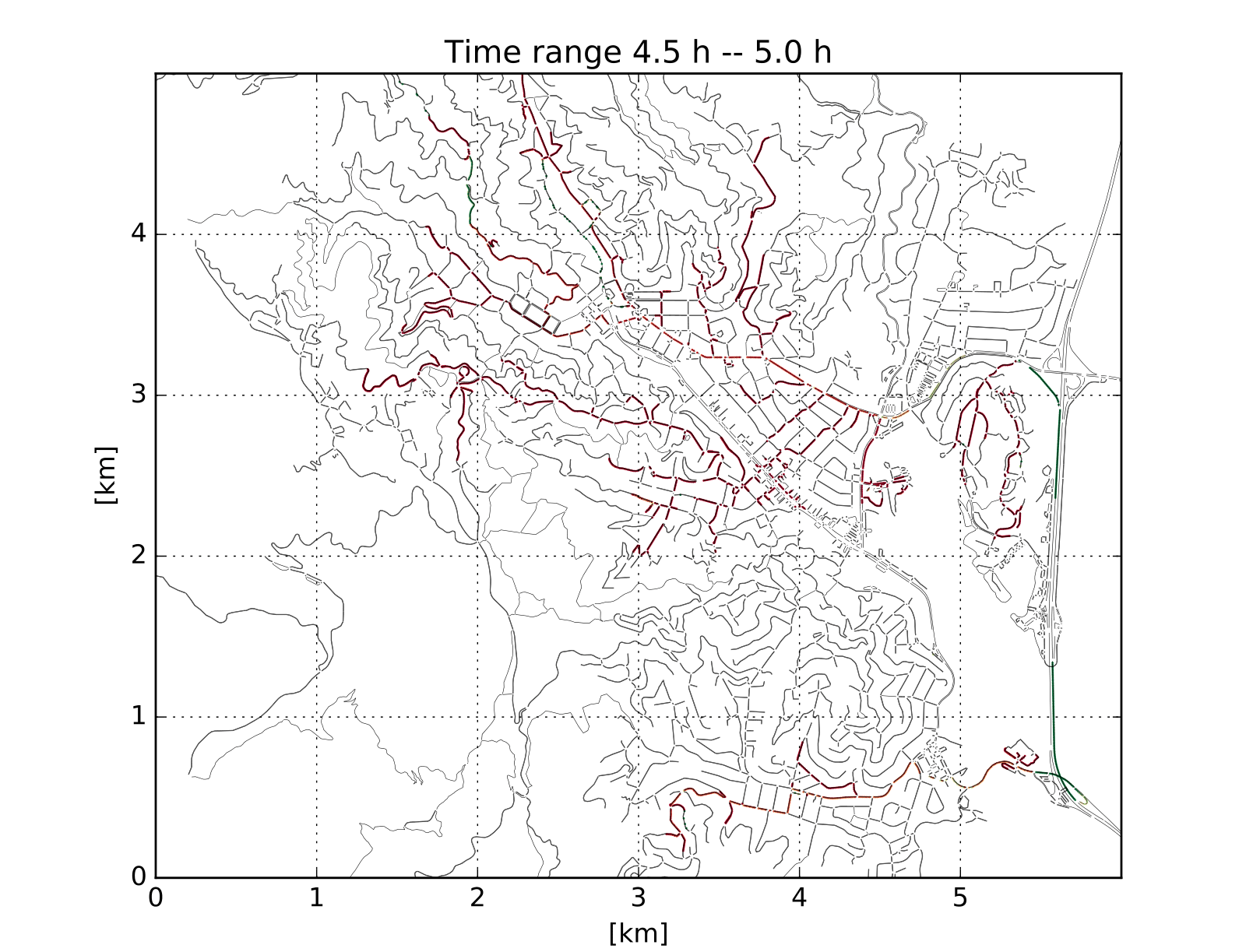}
    \caption{Scenario 8}
\end{subfigure}%
\begin{subfigure}[t]{0.5\textwidth}
    \centering
    \includegraphics[height=5cm]{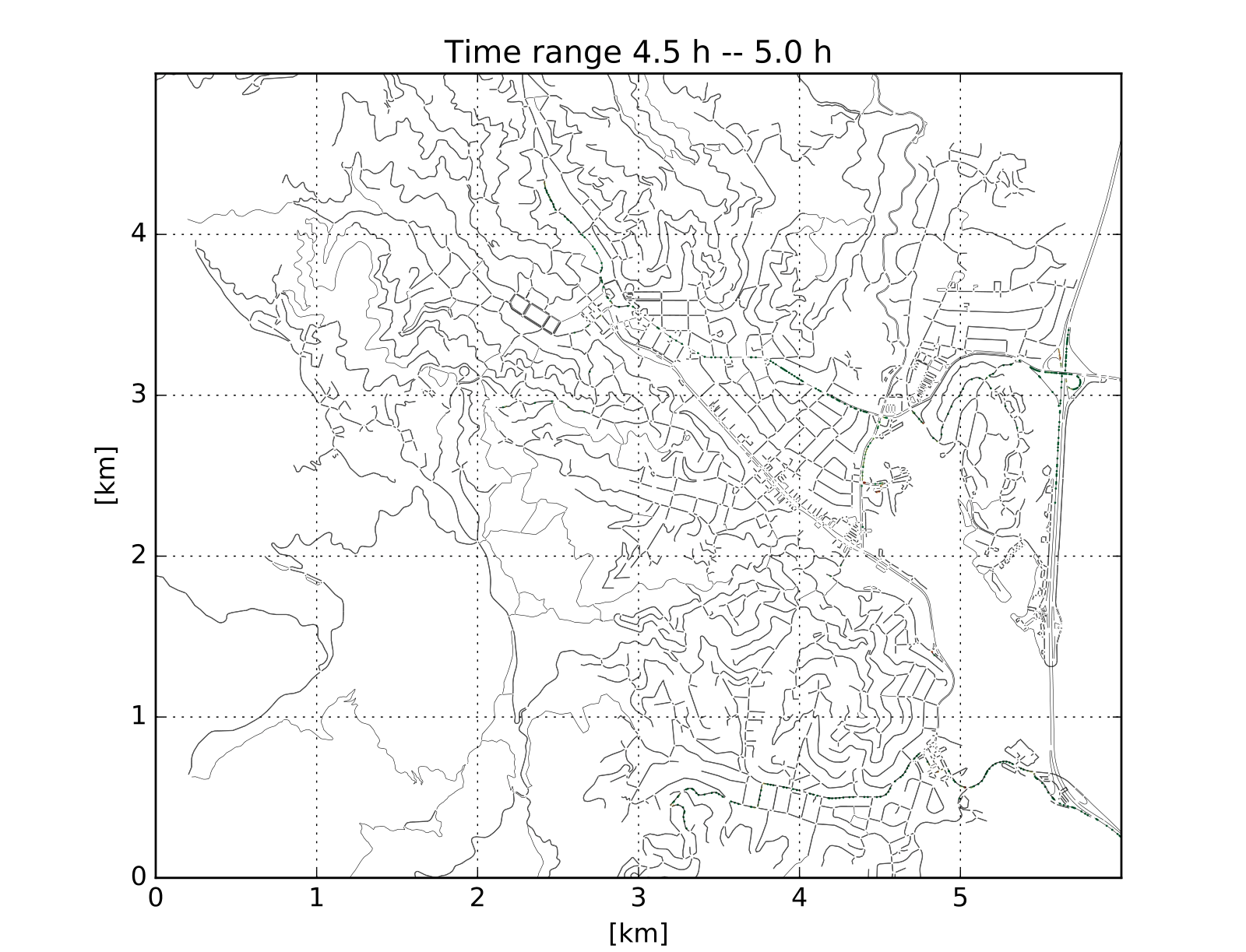}
    \caption{Scenario 10}
\end{subfigure}%

\caption{The speed maps for Mill Valley. The demands temporal distribution standard deviation is 0.5 h. (a, c, e, g) Scenario 8 in Table \ref{tab:scenarios} with shortest path routing and normal traffic management policy. (b, d, f, h) Scenario 10 with automatic routing and contraflow traffic management policy.}
\label{fig:speed_maps_millvalley}
\end{figure}

\paragraph{Speed maps} The speed maps for scenarios 1, 3, 8, and 10 are presented in Figure \ref{fig:speed_maps_paradise} and Figure \ref{fig:speed_maps_millvalley}. Scenarios 1 and 8 are the baseline scenarios, and 3 and 10 show the improvement using the evacuation traffic policy. The color track on a segment of road is the histogram of all recorded speeds at different position bins. The speed maps reflect both spatial and temporal dynamics of the traffic flow at a detailed level\footnote{The animation of the traffic simulation is easier to see in \textit{sumo-gui}. We provide necessary simulation files.}. They can be used to identify the bottlenecks of the network. 

\yuchen{Figure \ref{fig:speed_maps_paradise} (a, c, e, and g) are the speed maps for Paradise with the normal traffic management policy and shortest path routing (scenario 1 in table \ref{tab:scenarios}). Figure \ref{fig:speed_maps_paradise} (b, d, f, and h) are the speed maps for Paradise with the evacuation traffic management policy and automatic routing (scenario 3 in table \ref{tab:scenarios}). The standard deviation of the demands distribution in both cases is 0.7 hours. During the interval from 2 to 2.5 hours, the beginning of the evacuation and before the peak (Figure \ref{fig:speed_maps_paradise} (a)), a traffic jam emerges at the crossroad at Pentz Rd. and Pearson Rd. due to the stop sign. In comparison, a jam does not appear in the modified policy scenario (see Figure \ref{fig:speed_maps_paradise} (b)). During the demands peak interval between 3 and 3.5 hours (see Figure \ref{fig:speed_maps_paradise} (c)), the traffic jams in scenario 1 quickly expand to very large residential areas, which concentrate around the crossroads. In scenario 3, the speed map in Figure \ref{fig:speed_maps_paradise} (d) does not indicate clogged traffic along the arterial roads. The contraflow strategy significantly enlarges the capacity of the intersections. One hour after the peak during the interval between 4 and 4.5 hours (see Figure \ref{fig:speed_maps_paradise} (e)), the slow traffic flow in scenario 1 still remains almost the same, and the traffic jam is not alleviated in most areas. In scenario 3, since the evacuation policy does not have bottlenecks and the traffic does not accumulate during the peak interval, the traffic burden becomes lighter during this interval, with the condition remaining clear as before. Two hours after the peak time in the interval between 5 and 5.5 hours, in scenario 1, all of Pentz Rd. is still blocked. The middle part of Skyway Rd. still has slow areas. The traffic along Clark Rd. and the upstream portion of Skyway Rd. has been cleared. In scenario 3, the whole network is almost empty. It is worth noting that the traffic jam areas shown in scenario 1 roughly agree with the locations of fatalities and abandoned cars reported by \cite{john2018paradise}. The locations of fatalities concentrate in the southern area between Clark Rd. and Pentz Rd. with a few distributed along Pentz Rd. Many abandoned cars are located on east Pearson Rd., south Pentz Rd., Elliott Rd. and north Skyway Rd. The traffic jam in simulation roughly matches the locations of the abandoned cars except for the northern portion of Skyway Rd. The intersection issues in scenario 1 were also reported by eyewitness \cite{john2018paradise}. Quoting from the report, ``\textit{Broshears said he was surprised by how quickly intersections became a choke point. Traffic backed up on secondary roads so solidly that motorists were trapped on dead-end streets.}" Thus both simulation and what really was observed encourage us to improve the traffic evacuation problem by fixing the intersection as the highest priority. Our proposed policy shows promising results in tackling the problem. }

Figure \ref{fig:speed_maps_millvalley} (a, c, e, and g) are the speed maps for Mill Valley with normal traffic management policy and shortest path routing (scenario 8 in table \ref{tab:scenarios}). Figure \ref{fig:speed_maps_paradise} (b, d, f, and h) are the speed maps for Mill Valley with the new traffic management policy and automatic routing (scenario 10 in table \ref{tab:scenarios}). The standard deviation of the demands distribution in both cases is 0.5 hours. For the Mill Valley baseline scenario 8 in Figure \ref{fig:speed_maps_millvalley} (a), one hour before the peak comes, the network is quiet. There are no obvious traffic jams on the roads, the same as scenario 10 in Figure \ref{fig:speed_maps_millvalley} (b). Then, it is not surprising to see that when the demands peak arrives during the interval between 2.5 and 3 hours, E Blithedale Ave. quickly gets clogged by 4 traffic lights in the last 0.7 miles to U.S. 101 exits, see Figure \ref{fig:speed_maps_millvalley} (c). The traffic jam expands over all of E Blithedale Ave. and the connected secondary roads. It is mainly in the downtown area. The traffic jam also appears on Shoreline Hwy. As a comparison, the traffic condition in scenario 10 in Figure \ref{fig:speed_maps_millvalley} (d) is better. The arterial roads (paths $a, b, c$ in Figure \ref{fig:new_traffic_policies}) have relatively free flow conditions. But there are still few ``red" branches on roads because they have low priorities and they need to give the right-of-way to the main stream on the arterial roads. The new policy does not modify Shoreline Hwy., and the traffic condition in the south side of the city is similar to scenario 1. Keeping the arterial roads clear is a main advantage of the new traffic management policy since it will not result in accumulation of traffic in the road network, and following cars will not get trapped. This can be revealed in the post-peak interval between 3.5 and 4 hours. In Figure \ref{fig:speed_maps_millvalley} (e), the traffic jam extends from the downtown area to the peripheral regions. In scenario 10 (Figure \ref{fig:speed_maps_millvalley} (f)), traffic in part of the downtown is blocked, but not in the peripheral areas. \retouch{Evacuation of Mill Valley poses several considerations, since the city is located in a narrow valley, with the road network constrained by the landscape and topography.} The new policy enlarges the network capacity by adding a reversed road and a detoured local road, but still gets saturated during the peak. However, the jammed condition in scenario 10 does not last long. One hour later, the network is fully cleared as shown in Figure \ref{fig:speed_maps_paradise} (h). The traffic jam in scenario 8 in Figure \ref{fig:speed_maps_millvalley} (g) still remains in both the downtown area and the peripheral areas. There is no real case we can compare with, but the new evacuation policy shows remarkably improvement in the simulation.

\begin{figure}[H]
\centering
\begin{subfigure}[t]{0.5\textwidth}
    \centering
    \includegraphics[height=6cm]{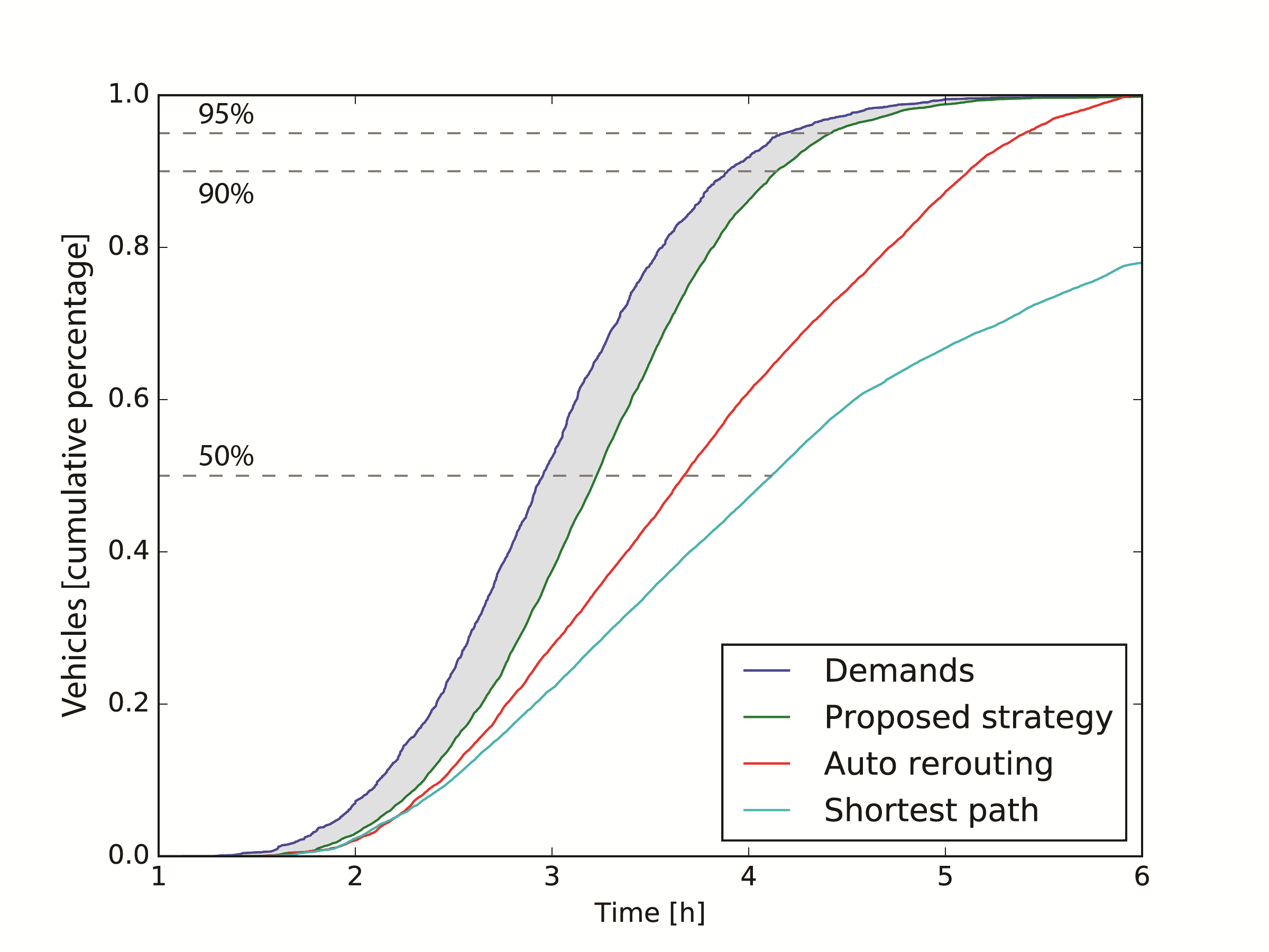}
    \caption{Paradise}
\end{subfigure}%
\begin{subfigure}[t]{0.5\textwidth}
    \centering
    \includegraphics[height=6cm]{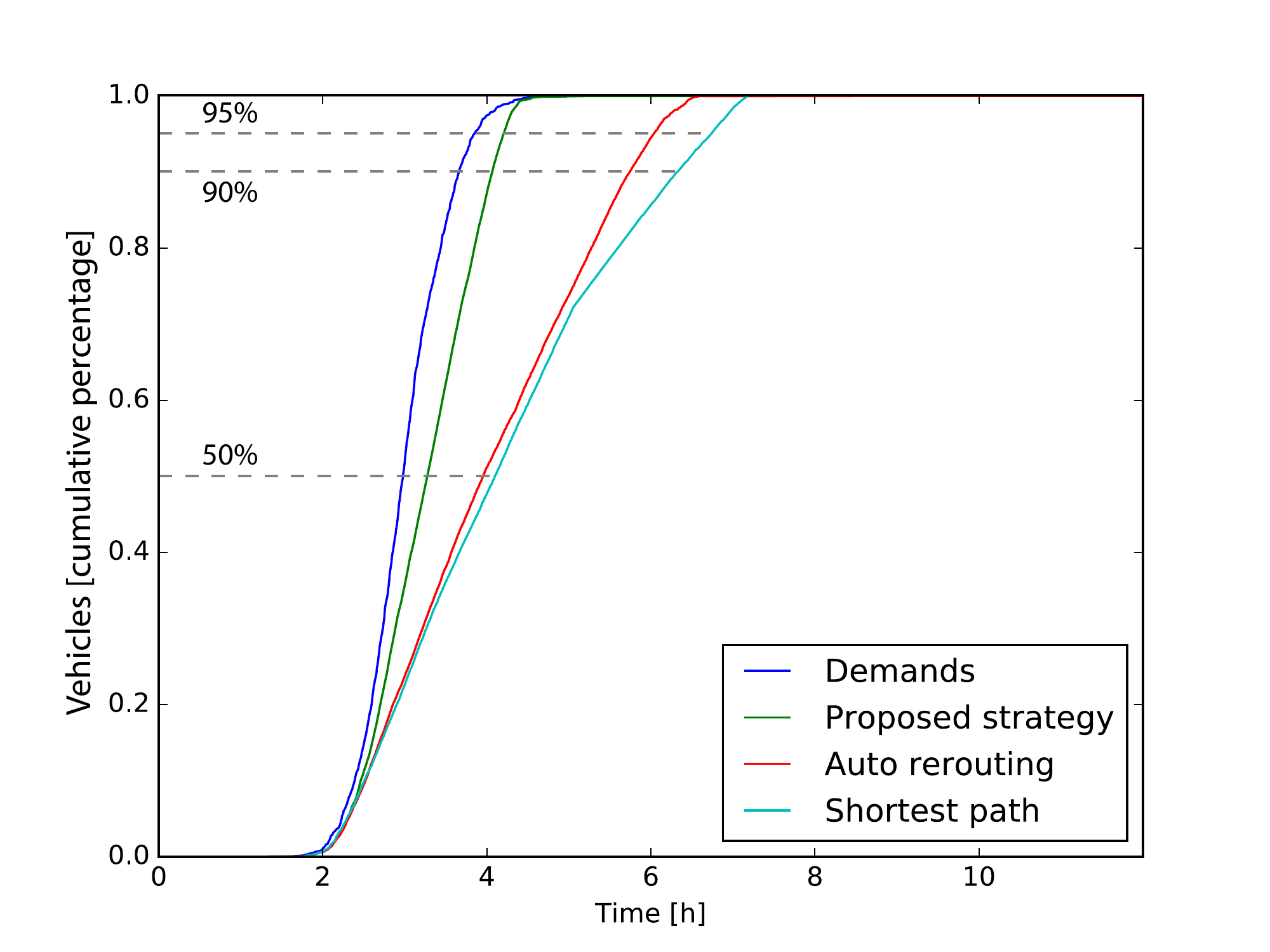}
    \caption{Mill Valley }
\end{subfigure}%

\caption{The demands-evacuation curves for scenarios 1, 2, 3, 8, 9, and 10 in table \ref{tab:scenarios}. (a) The demands-evacuation curves for several Paradise scenarios, including shortest path routing, automatic routing, and proposed new traffic management rules. The standard deviation for the demands is 0.7 h. The gap areas (defined in equation \ref{eq:gap_area}) are 0.25, 0.73, and 1.77. \yuchen{An example gap area between the demand curve and the proposed strategy is shown in the gray shadow.} (b) The demands-evacuation curves for several Mill Valley scenarios, including shortest path routing, automatic routing, and proposed new traffic management rules. The standard deviation for the demands is 0.5 h. The gap areas are 0.26, 1.05, and 1.24. }
\label{fig:demands_evacuation_curves}
\end{figure}

\paragraph{Demands-evacuation curves}
The speed maps present the details of the evacuation procedure, while the demands-evacuation curves depict the overall picture. Figure \ref{fig:demands_evacuation_curves} (a) summarizes the scenario 1, 2, and 3 in table \ref{tab:scenarios} with the same demands for Paradise but with different routing methods and traffic management policies. The standard deviation of the demands temporal distribution is 0.7 hours. The gap areas (defined in equation \ref{eq:gap_area}) are 0.25, 0.73, and 1.77. This implies that the new traffic management policy is around three times faster at evacuating the city than the baseline scenarios. The slope of the new scenario evacuation curve is almost in parallel with the demands curve, while the baseline scenarios' evacuation curves deviate significantly from the demands curve. This indicates that the network is saturated in the baseline scenarios and is not saturated in the new policy scenario. The latency between the 50\% demands and 50\% evacuations for the baseline scenarios is about 1 hour for the automatic routing method and 1.5 hours for the deterministic shortest path routing. The latency between the demands curve and evacuation curve is shorter at lower percentages and becomes longer at higher percentages. This is because the traffic jam is lighter at earlier stages and worse at later stages. As a comparison, the latency for the new scenario is about 0.36 hours at almost all percentages. The new policy is able to evacuate the vehicles at the same rate as the traffic generation rate. However, the baseline scenarios cannot accommodate the large volume of traffic, so more and more vehicles get stuck on the roads. This agrees with the speed map in Figure \ref{fig:speed_maps_paradise}. Figure \ref{fig:demands_evacuation_curves} (b) summarizes scenarios 8, 9 and 10 in table \ref{tab:scenarios} with the same demands for Mill Valley. The gap areas are 0.26, 1.05, 1.24. The standard deviation of the demands temporal distribution is 0.5 hours. The results are similar to those of Paradise. Note that there is a larger gap between the auto rerouting scenario and the deterministic scenario in Paradise, but such a gap is smaller in Mill Valley. Paradise has four exits while Mill Valley has only two exits. It is more important to balance the traffic load in Paradise, while in Mill Valley, there are not a lot of options.

\begin{figure}[H]
\centering
\begin{subfigure}[t]{0.5\textwidth}
    \centering
    \includegraphics[height=6cm]{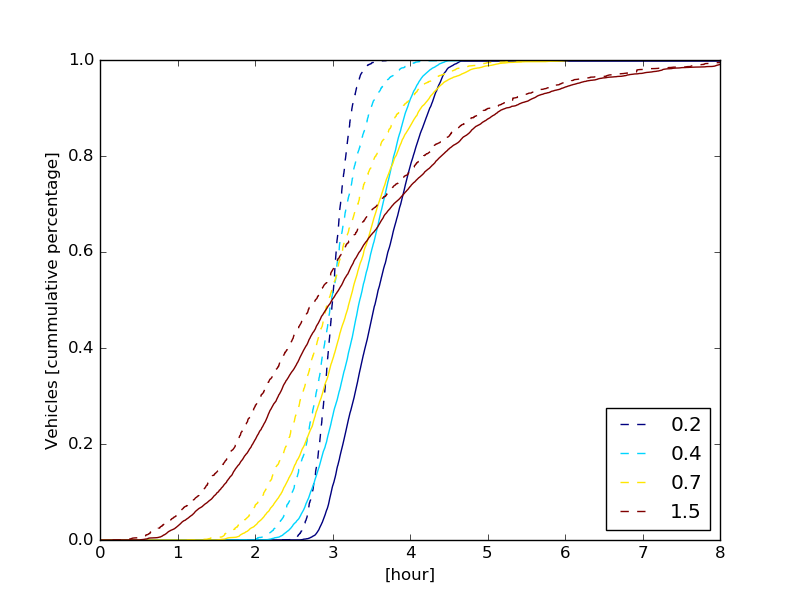}
    \caption{Paradise }
\end{subfigure}%
\begin{subfigure}[t]{0.5\textwidth}
    \centering
    \includegraphics[height=6cm]{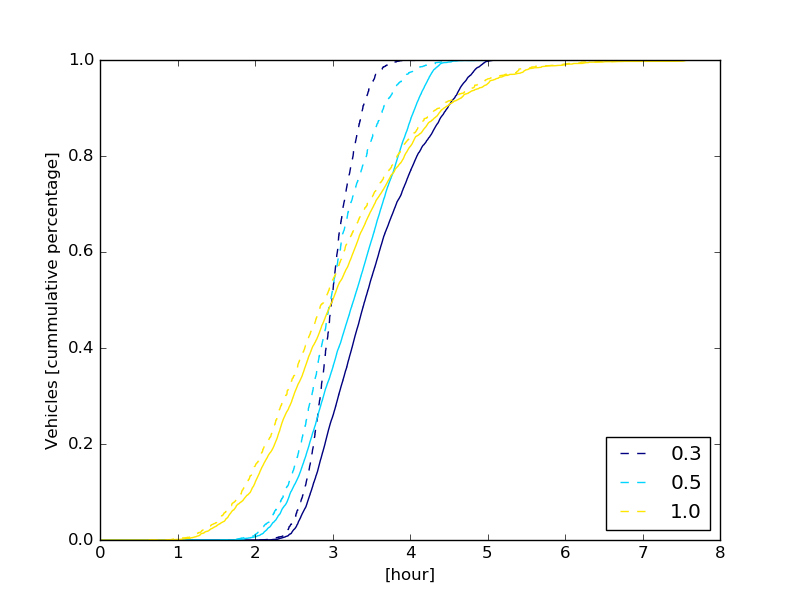}
    \caption{Mill Valley }
\end{subfigure}%

\caption{The relationship between the demands-evacuation curves and the concentration of the demands for scenarios 4 and 11 in table \ref{tab:scenarios}. (a) Paradise. The demands-evacuation curves for the new traffic management scenarios with different concentrations of the demands. The standard deviations are 0.2 h, 0.4 h, 1.0 h, and 1.5 h. The gap areas are 0.22, 0.25, 0.36, and 0.59. (b) Mill Valley. The demands-evacuation curves for the new traffic management scenarios with different concentrations of the demands. The standard deviations are 0.3 h, 0.5 h, and 1.0 h. The gap areas are 0.10, 0.26, and 0.50.  }
\label{fig:multi_demands_evacuation_curves}
\end{figure}

\paragraph{New traffic management policy with different demands concentration}
Next we study how the demands concentration affects evacuation for the new traffic management policy. Figure \ref{fig:multi_demands_evacuation_curves} summarizes scenarios 4 and 11 in table \ref{tab:scenarios}. It presents the evacuation efficiency with different demands ranging from more to less concentrated temporal distribution. These cases correspond to the no-notice and short-notice evacuations. For Paradise, the gap areas are 0.22, 0.25, 0.36, and 0.59 for demands standard deviations being 0.2, 0.4, 0.7, and 1.5. When the concentration is low, for example if the stranded deviation of the demands distribution is 0.7 hour or 1.5 hour, the evacuation curves remain parallel to the demands curve. If the demands concentration increases, the evacuation curves deviate from the demands curve. This means the network gets saturated, and it reflects the maximum capacity of the road network during an evacuation. The evacuation curves for standard deviations 0.4 hours and 0.2 hours are almost the same except for time shifts. For saturated cases, the evacuation efficiencies stay the same. So increasing the demands concentration will not help the evacuation but will cause more vehicles to enter and clog the roadways and potentially lead to more accidents. For less urgent evacuation, for example where it is feasible to generate demands modeled with a standard deviation larger than 0.7 hours, it is possible to implement part of the new policy to achieve the same performance. For more urgent evacuation which requires the standard deviation to be smaller than 0.4 hour, alerting all residents to leave in a very short time does not make the evacuation happen more quickly, but instead may cause mass panic and anxiety on the road. For Mill Valley, the gap areas are 0.10, 0.26, and 0.50 for demands standard deviations being 0.3, 0.5, and 1.0, and the results are similar to those of Paradise.

\begin{figure}[H]
\centering
\begin{subfigure}[t]{0.5\textwidth}
    \centering
    \includegraphics[height=6cm]{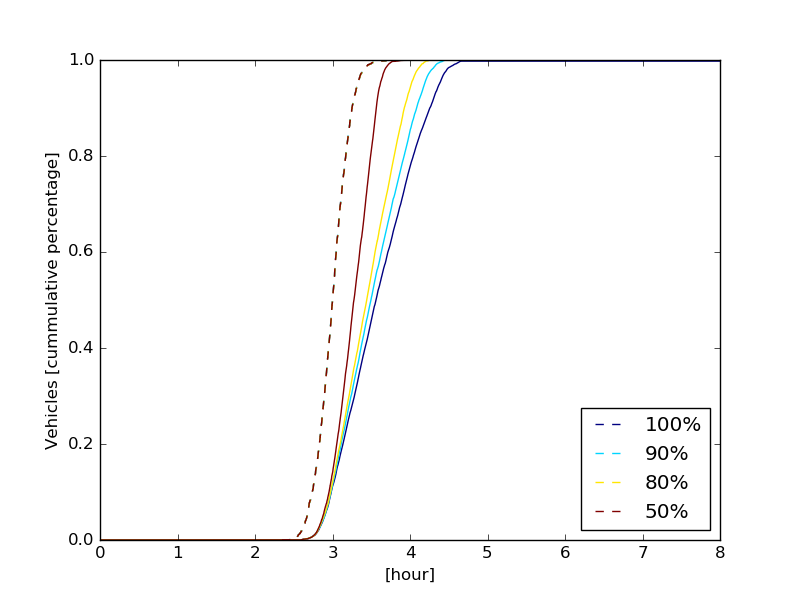}
    \caption{Paradise saturated }
\end{subfigure}%
\begin{subfigure}[t]{0.5\textwidth}
    \centering
    \includegraphics[height=6cm]{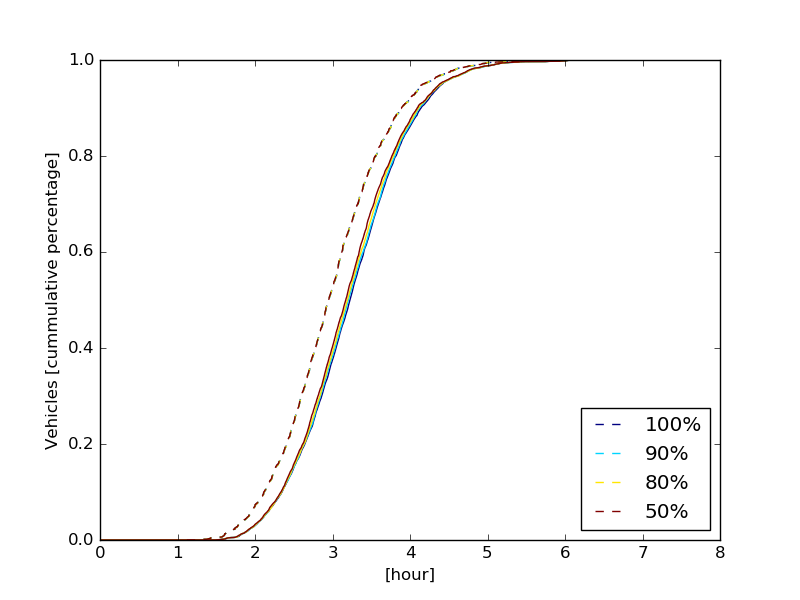}
    \caption{Paradise non-saturated }
\end{subfigure}%

\begin{subfigure}[t]{0.5\textwidth}
    \centering
    \includegraphics[height=6cm]{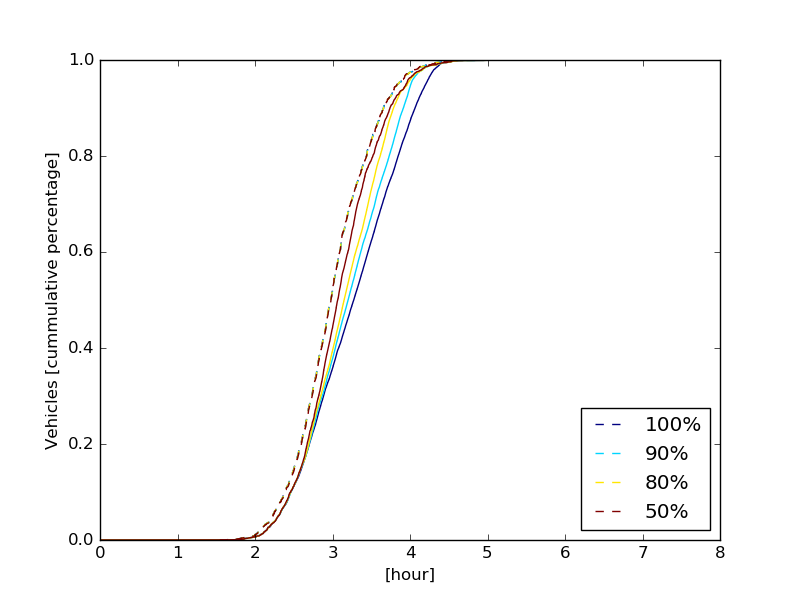}
    \caption{Mill Valley saturated }
\end{subfigure}%
\begin{subfigure}[t]{0.5\textwidth}
    \centering
    \includegraphics[height=6cm]{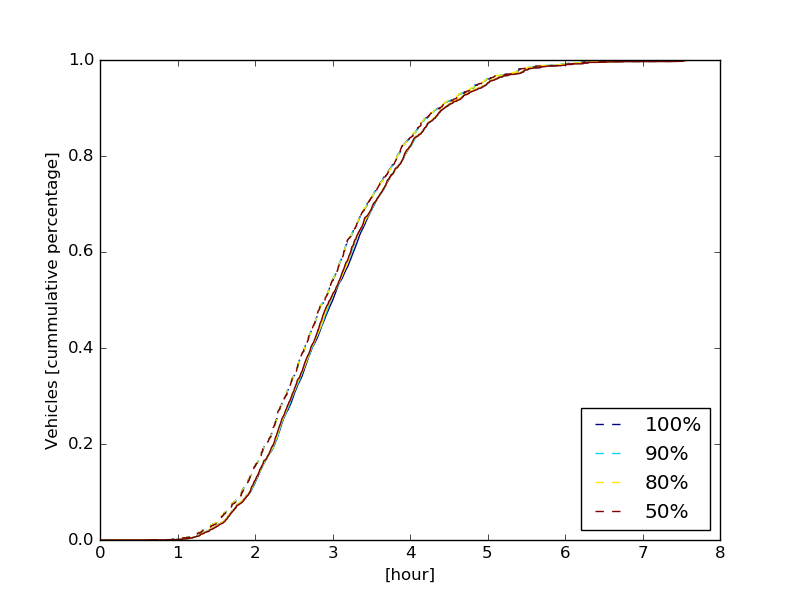}
    \caption{Mill Valley non-saturated }
\end{subfigure}%

\caption{The relationship between the demands-evacuation curves and the portion of the population for scenario 5, 6, 12, and 13 in table \ref{tab:scenarios}. (a) Paradise. The demands-evacuation curves for the new traffic management scenarios with different portions of the population. The portions are 100\%, 95\%, 90\%, and 50\% of the whole population. The standard deviation for the demands is 0.2, and the evacuation is saturated. The portion of 100\% is also shown in Figure \ref{fig:multi_demands_evacuation_curves}. The gap areas are 0.59, 0.52, 0.45, and 0.27 for the portion being 100\%, 95\%, 90\%, and 50\%. (b) Paradise. Similar to (a), The portions are 100\%, 95\%, 90\%, and 50\% of the entire population. The difference from (a) is that the standard deviation is 0.5 and the evacuation is not saturated.  The gap areas are 0.25, 0.24, 0.23, and 0.21 in the same order of the portions. (c) Mill Valley. The demands-evacuation curves for the new traffic management scenarios with different portions of the population. The portions are 100\%, 95\%, 90\%, and 50\% of the whole population. The standard deviation for the demands is 0.5, and the evacuation is saturated. The portion of 100\% is also shown in Figure \ref{fig:multi_demands_evacuation_curves}. The gap ares are 0.26, 0.19, 0.14, and 0.09.  (d) Mill Valley. Similar to (c), The portions are 100\%, 95\%, 90\%, and 50\% of the entire population. The difference from (b) is that the standard deviation is 0.7 and the evacuation is not saturated. The gap areas are 0.10, 0.09, 0.09, and 0.08.}
\label{fig:portion_demands_evacuation_curves}
\end{figure}

\paragraph{New traffic management policy with different portion of evacuated vehicles}
Figure \ref{fig:portion_demands_evacuation_curves} compares the evacuation efficiencies with different portions of evacuated vehicles for scenarios 5, 6, 12, and 13 in table \ref{tab:scenarios}. Figure \ref{fig:portion_demands_evacuation_curves} (a) shows the cases with high demands concentration, the standard deviation is 0.2 hours. The evacuation is saturated with 100\% vehicles in this case as shown before in Figure \ref{fig:multi_demands_evacuation_curves} (a). If the total number of vehicles is reduced, it helps to increase the evacuation. The gap areas are 0.59, 0.52, 0.44, and 0.27. The network remains saturated at 80\%-90\% of the total vehicles. However, if the evacuation is less concentrated and the network is not saturated, reducing the total number of vehicles does not help as shown in Figure \ref{fig:portion_demands_evacuation_curves} (b), where the gap ares are 0.25, 0.24, 0.23, and 0.21. For Mill Valley, the gap areas in Figure  \ref{fig:portion_demands_evacuation_curves} (c) for saturated cases are 0.26, 0.19, 0.14, and 0.09. The gap area for non-saturated cases in Figure \ref{fig:portion_demands_evacuation_curves} (d) are 0.10, 0.090, 0.09, and 0.08. The results are similar to those of Paradise.

\begin{figure}[H]
\centering
\begin{subfigure}[t]{0.5\textwidth}
    \centering
    \includegraphics[height=6cm]{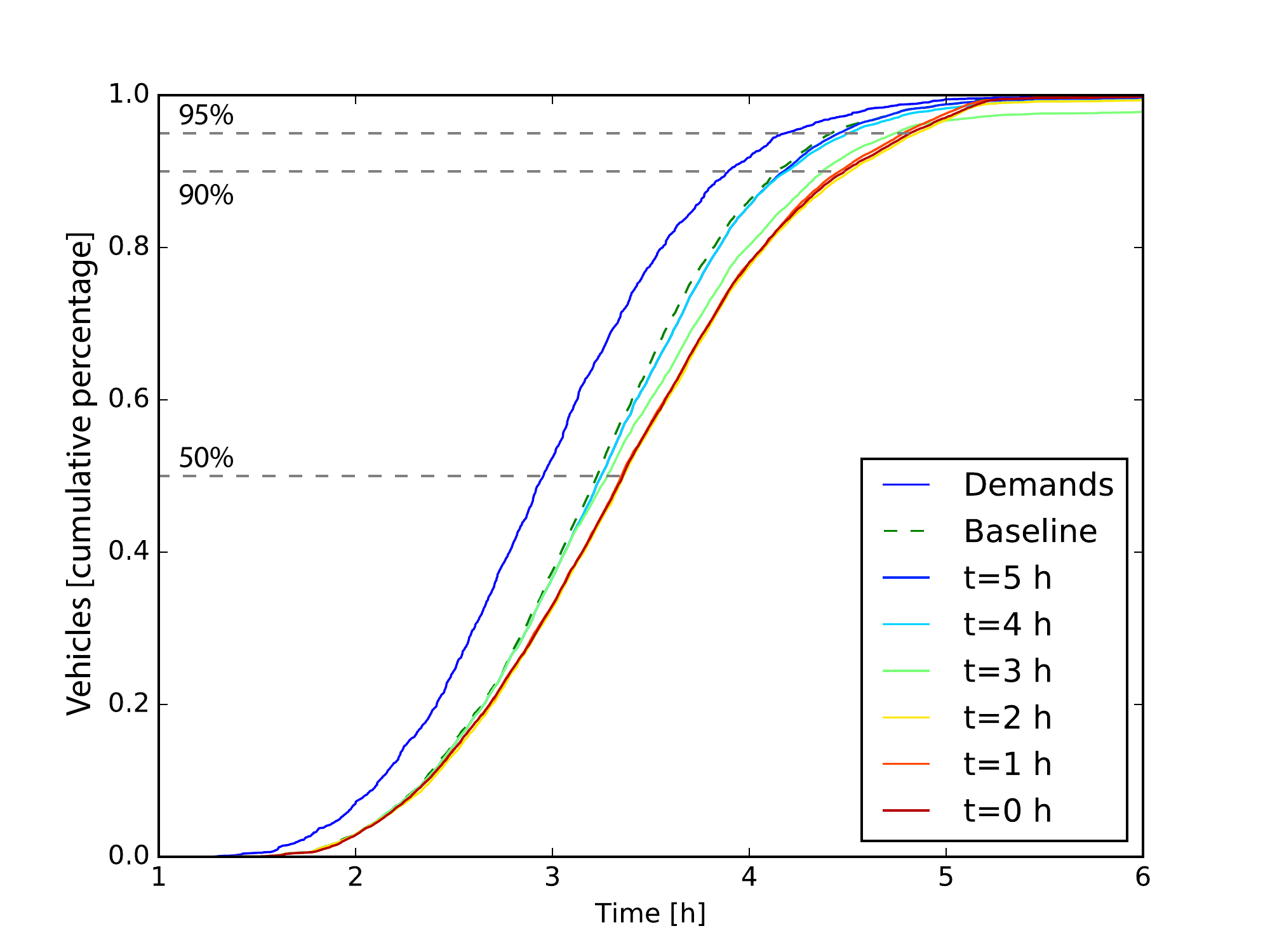}
    \caption{Paradise saturated }
\end{subfigure}%
\begin{subfigure}[t]{0.5\textwidth}
    \centering
    \includegraphics[height=6cm]{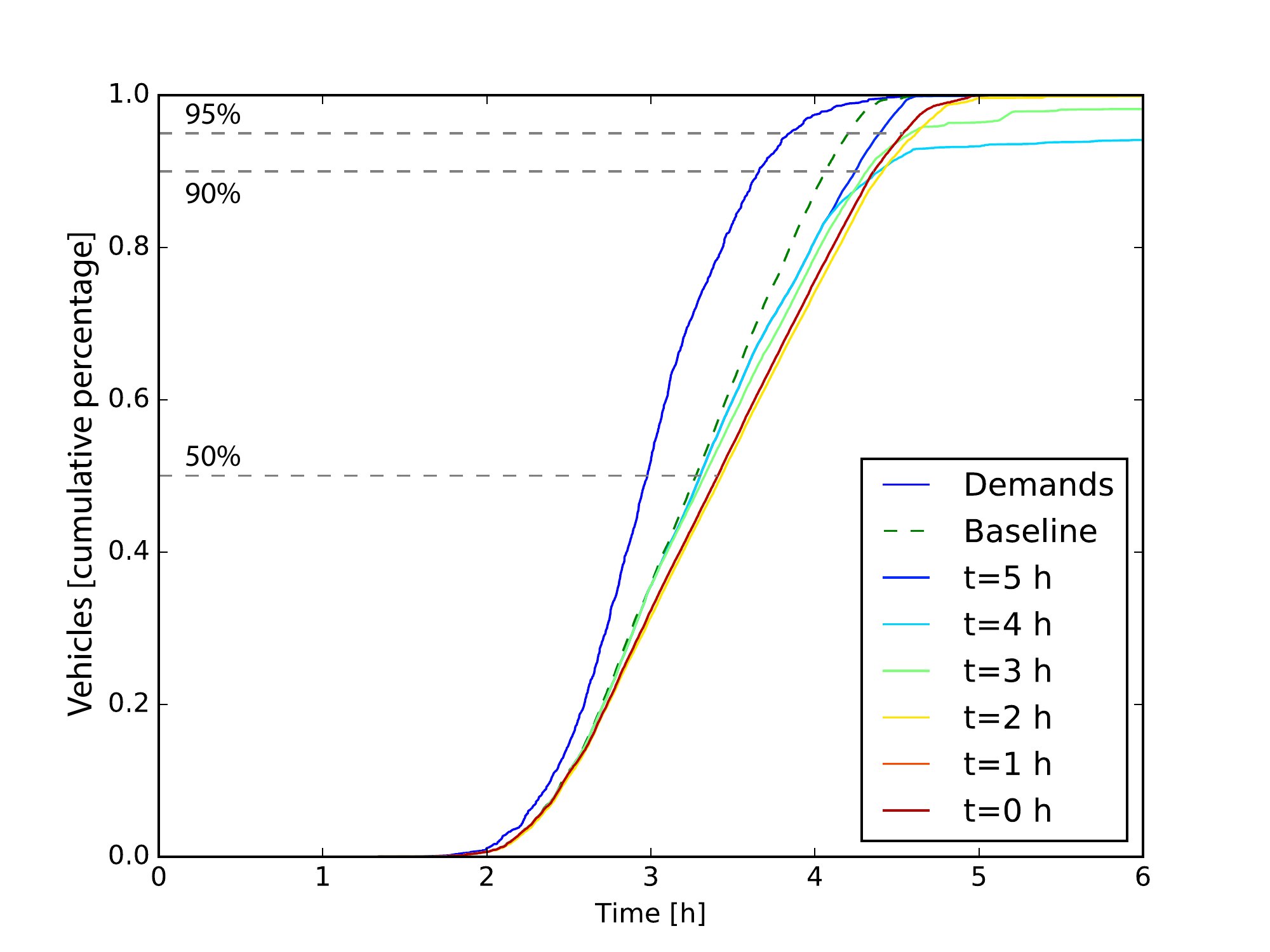}
    \caption{Mill Valley saturated }
\end{subfigure}%

\caption{Demands-evacuation curves under road closures during evacuation. These scenarios mimic the cases where the traffic management policies or the road network are disturbed by uncontrolled factors, such as blocking of the roads due to natural disaster impacts, traffic accidents, or traffic gridlock. (a) Paradise. It was reported that the fire had already reached Pentz Rd. during the evacuation of the Camp Fire. In our simulations, Pentz Rd. is closed at various times after the start of the fire. The gap areas are 0.25, 0.28, 0.30, 0.46, 0.44, 0.39, and 0.40 for baseline and closure times at 5, 4, 3, 2, 1, and 0 h. (b) Mill Valley. Mill Valley has not experienced a large fire recently, but the main exit road is closed in the simulation to evaluate the robustness of the evacuation plan. In the simulation, E Blithedale Ave. is closed at different times after the start of the fire. The gap areas are 0.26, 0.32, 0.67, 0.47, 0.44, 0.41, and 0.41 for baseline and closure times at 5, 4, 3, 2, 1, and 0 h.}
\label{fig:disturbance_demands_evacuation_curves}
\end{figure}

\yuchen{\paragraph{Robustness of the evacuation traffic management policy}
Finally, we evaluate the robustness of the new evacuation policy in part by perturbing the road network. Figure \ref{fig:disturbance_demands_evacuation_curves} presents the evaluation efficiency with the perturbation of the network for scenarios 7 and 14 in table \ref{tab:scenarios}. For Paradise, Pentz Rd. is blocked at different times, from the beginning the of the evacuation to five hours later. The gap areas are 0.25, 0.28, 0.30, 0.46, 0.44, 0.39, and 0.40 in the order of the legend in Figure \ref{fig:disturbance_demands_evacuation_curves} (a). If the road is blocked earlier, the network is limited with a reduced capacity for a longer period of time, and the gap area is larger and evacuation curve stays farther from the demands curve. Paradise has four exits, so blocking one of them for one segment of time does not influence the overall performance that much. If the road is blocked at the peak time (hour three) of the demands, more vehicles get stuck on the road and have difficulty in turning back to other exits. \\
For Mill Valley, path $a$ in Figure \ref{fig:new_traffic_policies} (d) is blocked at different times. The gap areas are 0.26, 0.32, 0.67, 0.47, 0.44, 0.41, and 0.41 in the order of the legend in Figure \ref{fig:disturbance_demands_evacuation_curves} (b). The conclusion is similar to that of Paradise.
}

\section{Discussion}\label{sec:discussion}

\paragraph{Maps} OpenStreetMaps provides detailed maps about many cities around the world, which is suitable for most users. As for the correction of the details, we give a list of possible errors in section \ref{subsec:methods_maps}  according to our experience. However, the types of errors may not be limited to those. As for very large scale maps, verifying the map may be more involved. We suggest that users inject more specified and localized traffic flows to examine the road and lane connections, instead of approaching the problem at the full map scale, as the computational complexity of the simulation may hinder speed of evaluation.

\paragraph{Demands} The demands are modeled using spatial and temporal distributions. We assume the demands are uniformly distributed in the city based on the observation that the majority of both Paradise and Mill Valley consists of residential areas with relatively uniformly distributed single-family houses, and the vehicles on each road are generated independently. However, this assumption should be refined for other cities with significantly uneven distribution of population and vehicles. For the temporal distribution, since the exact ground truth of the evacuation demands is not available for historical evacuation events, and it is impossible to predict the demands for urban evacuation planning with so many uncertain factors partially listed in \cite{murray2013evacuation}, we have to compromise on the temporal accuracy. Instead, the demands models with parameters in a large range are presented, capturing the range of demands that could be experienced in a real event. The interaction between spatial and temporal distributions is simplified by assuming they are independent: in other words, when a vehicle is generated does not depend on where it is generated. The validity of this spatial and temporal independence assumption relies on the alert system of the city. If the residents in one area learn about the event significantly earlier than those in another area, for example if the residents are informed by police officers driving from one region to another, the assumption should be refined, a direction that we are actively pursuing in follow-up work.

Here we propose a more advanced model for the demands generation using spatial and temporal point processes. This model has been successfully applied in fields as diverse neuroscience \cite{pillow2008spatio,kass2014analysis}, social networks \cite{kobayashi2016tideh}, seismology \cite{ogata1998space}, and finance \cite{hawkes2018hawkes}. It can used to fit the demands data if it is available, or to incorporate prior domain knowledge. Point processes describe stochastic discrete events, such as the generation of cars, on a continuous space, such as time line or a Cartesian plane. How likely a certain number of vehicles are generated is related to the underlying event rate function $\lambda(t, s)$ of time $t$ and spatial position $s$. The log-likelihood of generating data $x_1 = (t_1, s_1), ..., x_n = (t_n, s_n)$ in a time range and spatial range $T \times S$ is given below \cite[chapter 7]{daley2003basic},
\begin{equation}
\log L(x_1, ..., x_n) = \sum_{i=1}^n \log \lambda(t_i, s_i) - \int_{T \times S} \lambda(t, s) \mathrm{d} t \mathrm{d} s
\end{equation}
There are several special cases of the model above.
\begin{enumerate}
\item If the process is spatially homogeneous, then the function $\lambda(t, s)$ is a constant of $s$, which can be simplified into $\lambda(t)$. The the number of vehicles of the whole area $S$ or a subarea is an inhomogeneous a Poisson process. For any two areas with the same size, the number of generated vehicles follows the same inhomogeneous Poisson process.
\item If the process is temporally homogeneous, then the function $\lambda(t, s)$ is a constant of $t$, which can be simplified into $\lambda(s)$. The the number of vehicles of the whole area $S$ or a subarea is a homogeneous Poisson process.
\item If the process is both temporally and spatially homogeneous, then the function $\lambda(t, s)$ is a constant of $t$ and $s$, which can be simplified into $\lambda$. The the number of vehicles of the whole area $S$ or a subarea is simply a Poisson process. For any two areas with the same size, the number of generated vehicles follows the same Poisson process.
\item To simplify the model fitting and data generation, a map can be cut into blocks. Within each block, the process can be assumed to be a spatially homogeneous process. In this way, a continuous problem can be divided into a discrete problem. If the rate of generation is consistent in the time dimension, it can further be assumed to be a temporally homogeneous process. Thus fitting a whole area is equivalent to fitting several Poisson processes.
\item The model can be improved to be history dependent, i.e. generating a vehicle affects the generation rate in the future. Then $\lambda(t, s)$ becomes $\lambda(t, s | \mathcal{H}_t)$. $\mathcal{H}_t$ includes the data points $x_1, x_2,...$ up to time $t$. For example, in a block $A$, if the process is spatially homogeneous, then the number of vehicles in the area only depends on time. If process is a time-dependent or a self-excited process, then the underlying function is,
    \begin{equation}
    \lambda_A(t | \mathcal{H}_t) = \mu(t) + \sum_{(\tau_i, s_i) \in \mathcal{H}_t} \nu_A (t - \tau_i),
    \end{equation}
with $\mu(t)$ representing the time-varying baseline. It is independent of historical events. The parameter $\nu$ describes how history events affect the event rate function $\lambda$ at time $t$. In another word, the probability of observing an event at time $t$ depends on the timing of history events. For example, if drivers call their neighbors when they leave, the probability of someone leaving depends on the departure of his or her neighbors. If the underlying probability also depends on other information of an event, we refer readers to \textit{marked point process} for further reference. Since the underlying event rate function $\lambda$ itself is stochastic, the process like the above equation is also called a \textit{doubly stochastic process}.
\item The model can be refined to be spatially dependent, i.e. generating a vehicle in block $A$ affects the generation rate in block $B$. For example, the vehicle generation of block $B$ interacts with blocks $A_1,...,A_m$. If the process is a multivariate Hawkes process, then the underlying function for block $B$ is,
    \begin{equation}
    \lambda_B(t | \mathcal{H}_t) = \mu(t) 
    + \sum_{(\tau^B_i, s^B_i) \in \mathcal{H}^B_t} \nu_B (t - \tau^B_i)
    + \sum_{j=1}^{m} \sum_{(\tau^j_i, s^j_i) \in \mathcal{H}^j_t} \nu_{A_j} (t - \tau^j_i)
    \end{equation}
Similar to the last model, the underlying function $\lambda$ for area $B$ depends on the history of itself, besides that, it also depends on the events from other areas. It implies that whether a driver will leave a city depends on the status of other drivers in other regions. For example, if a driver $a$ in area $A$ has social connections with a driver $b$ in area $B$, $a$ may call $b$ when he/she leaves. 
\end{enumerate}
These models can be fitted by maximizing the likelihood function. It can also be treated as a generative model to create demands for simulation study. This framework is flexible with great potential. We provide preliminary tools for data fitting and data generation.

In some cities, the traffic flows are monitored by the induction loops, and so the demands can be reconstructed using the traffic flow at certain locations. We refer to \cite{bieker2015traffic} for more details.

When the traffic jam is extremely dense, the simulation can be less realistic because of the increasing probability of \textit{teleporting}, which handles the dead-lock by temporarily removing the car and adding it back later. Usually, though, in real scenarios the drivers can coordinate with each other to handle this situation with more or less aggressive behavior. In our study we do not increase the demands temporal distribution concentration to a very high level if the network is already saturated.

\paragraph{Routing} As introduced in section \ref{sec:routing}, the dynamic routing is performed independently for individuals, and the shortest path is calculated for the sake of each driver. However, this may not be a globally optimal strategy. A classic example is the Braess’s paradox with the ``diamond" structure \cite{rapoport2009choice}, where every driver chooses the fastest road, but the whole network gets slower. The network can experience better throughput when certain links are removed, which requires less instead of more resources. In our case studies, the maps are almost tree-like or forest-like, so there are not many diamond structures. However, in grid street maps, reducing or blocking several inner edges can potentially help to improve the overall traffic performance and reduce traffic flow conflicts. 

For scenarios 7 and 14 in table \ref{tab:scenarios} with network disruption, we apply dynamic routing to let some drivers change their original plans. In practice, letting drivers know the global information about the network on some level is critical, as they will not head to a dead-end road.

\paragraph{Traffic management policy} The proposed traffic management policy is simple and consistent, and in particular is easy for drivers to follow and does not highly rely on complex control systems, driver behavior, or supervision. Thus, adapting to the new policy requires minimal public education. Reducing the manpower needed is another reason to design simple and consistent policies. The policy is still applicable during power outages as it does not depend on the actuation of traffic lights. To avoid confusion about taking reversed roads, some guidance such as signs or police instruction is required. As for reversing a lane in the arterial road for the input traffic, temporary physical barriers may need to be placed.

\retouch{The traffic management policies proposed herein present rules of thumb to improve evacuation efficacy. However, potential policies are not constrained to the ones proposed, and in general depend on city-specific cases where standard or normative rules may not apply. For example, we do not consider policies requiring extra infrastructure. Previous works \cite{jha2004emergency,kumar2016optimizing} improve the evacuation performance through infrastructure upgrades, such as the expansion of the number of lanes and the construction of new roads. With such investments, residents can also benefit from the extra infrastructure in their daily lives. On the other hand, Paradise already has four arterial roads and speed maps of the new policy do not indicate larger traffic jams, so adding new roads probably cannot yield significant further improvement. For Mill Valley, additional road infrastructure could theoretically yield better outcomes given the current limited number of exit roads, and enhance the robustness of the policy, but physical constraints arising from the geography and topography limit this possibility.}

Our results show that when the demands are highly concentrated and the network is saturated by the traffic flow, reducing the number of vehicles can improve the evacuation efficiency. We point out a caveat in implementation and public education. Figure \ref{fig:portion_demands_evacuation_curves} (a) or (c) shows that if half of the vehicles are reduced, the total evacuation time can decrease by around one hour. This strategy is optimal on the whole population level: abandoning 50\% of vehicles (twelve thousands cars for Paradise and six thousands cars in Mill Valley) can save one hour in total. On average, however, it may only result in savings around half a second for an individual household leaving one car behind. If the evacuation is not urgent and the traffic is not saturated, the time saved is smaller. So it may be difficult justifying leaving cars behind as the time gain for an individual family could be negligible compared with the loss of property. If the residents are not motivated to leave cars behind, the city should not rely on reducing cars to improve the evacuation. This example is analogous to the Prisoner's dilemma in Game theory, where if each individual optimizes for their own outcome, the total outcome can be worse for everyone.

It is important for officials to make plans about when to change the normal traffic policy to an emergency evacuation policy. A late order becomes hard to execute, and have severe consequences \cite{john2018paradise}. This may also involve estimation about whether the fire will spread to the urban area and how fast it will move. Fire dynamics and simulation is outside of the scope of this paper.

\retouch{We build new traffic management policies on existing transportation infrastructure. Adding new roads, bridges or extra lanes could also be an option for city evacuation plans, especially to handle bottlenecks. For example, even though contraflow measures and traffic detours significantly increase the capacity of the network, Mill Valley's E Blithedale Ave. still remains a problem. Theoretically, further improvement could be had by widening the road, although physical constraints limit the feasibility of such interventions.  Our simulation pipeline can nonetheless serve as guidance for the impact of such hypothetical changes.}

\retouch{We design, test, and tune the proposed traffic management policies by manually designing the rules. This approach works well for cities like Paradise and Mill Valley. If the region of interest is much larger than these cities and the networks are more complicated, it will become impractical to do such analysis by hand. Programmatic tools need to be developed for these applications. On the other hand, policy design and analysis depends on tractable case studies. We do not expect to apply the same strategy for every city. Thus, programmatic tools can help with the work, but cannot obviate the need for careful calibration.}

\paragraph{Robustness of the policies}
We partially evaluate the robustness of the new evacuation traffic management policies by blocking one of the arterial roads in Paradise, and one arterial road in Mill Valley. A more thorough study can be done to evaluate the consequences of the network disruption or road capacity reduction. by blocking all possible arterial roads at different positions. These studies in turn add more constraints to the design of the traffic management policies. We will address this in future work.

\paragraph{Communication system}
One challenge for the whole city evacuation is the alarm system. Paradise had advertised its warning system which can promote “pack and go” preparations, and had included fire precautions into public construction project building codes. However, in practice, the warning system failed to reach more than a third of even the minority who enrolled \cite{paradise2018timeline}. In principle, alarm systems can also potentially be used to inform about fire and traffic conditions, so an evacuation can be more organized. 

\paragraph{External factors}
In section \ref{subsec:scenarios_demands}, we assume there is no background traffic on exit roads. This assumption is valid as long as the background traffic does not block the outgoing traffic flow. However, it can be a problem if there are traffic jams on the exit roads. It is possible to study the influence of the background traffic by injecting traffic flows with varying densities. Those external factors do not belong to city management functions, so they are not included in this work. It is a reminder that it is better to clear the main exits during the evacuation. 


\section{Conclusions. } \label{sec:conclusion}
In this paper, we provide a traffic simulation pipeline to study the urban area evacuation problem. With this pipeline, we are able to estimate the evacuation time, to identify the bottlenecks, and further to test existing and proposed evacuation traffic management policies and their robustness. Two case studies are presented using the pipeline. The contraflow method-based new evacuation policy significantly outperforms the normal traffic management policies. The temporal concentration of the demands influences the evacuation. In general, if the demands have higher concentration, the evacuation is faster. But if the network is saturated, increasing the demand concentration does not change the evacuation efficiency. In the saturated case, decreasing the portion of the population being evacuated alleviates the traffic jam. However, it does not help if the network is not saturated. The microscopic traffic simulation platform provides an opportunity to examine many details of the evacuation procedure, such as maps, routing algorithms, demands, and traffic management policies. We will publish the code and associated files, and hope other researchers and city managers can benefit from it for their further studies and city emergency preparedness.

\section{Acknowledgements}
We thank Mill Valley officials for useful discussions. We also thank Francois Belletti for his suggestions and draft review.




\end{document}